\documentclass[11pt]{article}

\usepackage[preprint]{acl}

\usepackage{times}
\usepackage{latexsym}

\usepackage[T1]{fontenc}

\usepackage[utf8]{inputenc}

\usepackage{microtype}

\usepackage{inconsolata}

\usepackage{graphicx}

\title{\system: A Fact Verification Benchmark over Large Structured Data}

\author{
 \textbf{Michael Theologitis\textsuperscript{1}},
 \textbf{Preetam Prabhu Srikar Dammu\textsuperscript{1}},
 \textbf{Chirag Shah\textsuperscript{1}},
 \textbf{Dan Suciu\textsuperscript{1}}
 \\
  \textsuperscript{1}University of Washington, \\
  Seattle, WA, USA \\
  \texttt{mthe@cs.washington.edu}
}

\usepackage{xcolor}
\newcommand{\gray}[1]{\textcolor{gray}{\footnotesize #1}}
\usepackage{booktabs}
\usepackage{multirow}
\usepackage{makecell}
\usepackage[cachedir=.]{minted}
\usepackage{fancyvrb}
\usepackage{pifont} 
\usepackage{enumitem}  
\usepackage{lipsum} 
\usepackage{arydshln} 

\newcommand\eat[1]{}
\usepackage{amssymb}  
\usepackage{amsmath} 
\usepackage{placeins} 
\usepackage{tikz}  
\newcommand*\circled[1]{\tikz[baseline=(char.base)]{
            \node[shape=circle,draw,inner sep=0.5pt] (char) {#1};}} 
\definecolor{greenish}{RGB}{102,204,0}
\usepackage{textcomp} 
\usepackage{xspace}
\usepackage{xcolor}

\definecolor{e}{RGB}{34,139,34}      
\definecolor{c}{RGB}{220,20,60}  
\definecolor{nei}{RGB}{12,64,236}          

\usepackage[colorinlistoftodos]{todonotes}

\usepackage{xspace}

\newcommand{\nop}[1]{}

\newcommand{\system}{\textsc{ClaimDB}\xspace}

\begin{document}
\maketitle 
\begin{abstract}
Real-world fact-checking often involves verifying claims grounded in structured data at scale. Despite substantial progress in fact-verification benchmarks, this setting remains largely underexplored. In this work, we introduce \system, a  fact-verification benchmark where the evidence for claims is derived from compositions of millions of records and multiple tables. \system consists of 80 unique real-life databases covering a wide range of domains, from governance and healthcare to media, education and the natural sciences. At this scale, verification approaches that rely on ``reading'' the evidence break down, forcing a timely shift toward reasoning in \textit{executable} programs. We conduct extensive experiments with 30 state-of-the-art proprietary and open-source (below 70B) LLMs and find that more than half score below 55\% accuracy. Our analysis also reveals that both closed- and open-source models struggle with \textit{abstention}---the ability to admit that there is no evidence to decide---raising doubts about their reliability in high-stakes data analysis tasks. We release the benchmark, code, and the LLM leaderboard at \url{https://claimdb.github.io}.
\end{abstract}

\section{Introduction}

Claims based on large-scale structured data are everywhere. They effectively drive and justify the most important decisions of our times. For example, Joe Biden stated on August 29, 2023 in a speech at the White House that the U.S. inflation \textit{``is now down close to 3, the lowest among the world’s leading economies.''}~\citeyearpar{WRAL-2022-biden} at the time of the most aggressive cycle of interest-rate hikes in decades~\cite{PBS-2022-inflation-largest-decades}. Similarly, two years later, on August 11, 2025, Donald Trump stated that \textit{``Washington, D.C., has 41 homicides per 100,000 people, No. 1 that we can find anywhere in the world.''}~\cite{WhiteHouse-2025-Donald-Trump} to justify the historic decision and deployment of the national guard at the U.S. capital (announced later in the same speech). 

In both cases, these claims are summaries of large, official, publicly available structured datasets. Biden's inflation claim can be verified against consumer price indices (CPI) published by the Bureau of Labor Statistics in the form of approximately ten separate Excel tables~\cite{BLS-Data}. On the other hand, Trump's crime claim can be fact-checked against crime records released by the Metropolitan Police Department of D.C. in the form of a single CSV file containing crime incidents~\cite{DC-Crime-Data}.

Despite the central role of structured data in real-world decision-making, fact-verification research has largely focused elsewhere. Prior work has made important progress on evidence \emph{grounded} in modalities ranging from text and documents to Wikipedia tables, info-boxes, and small relational tables (e.g.,~\citealp{DBLP:conf/iclr/ChenWCZWLZW20, DBLP:conf/nips/SchlichtkrullG023}). These settings have enabled significant advances in fact-checking (e.g., ~\citealt{DBLP:conf/iclr/BazagaLM24a}), but they share a common simplifying assumption: evidence is \emph{small}---in fact, it is small enough to fit within an LLM's context window. As Biden and Trump's examples showcase, this assumption does not hold for many high-stakes, real-world claims.

In this work, we propose \system, a fact-verification benchmark in which claims are grounded in evidence deliberately composed from multiple tables and millions of rows. Successful verification in \system implicitly requires some form of \textit{executable} reasoning ~\cite{DBLP:journals/tmlr/ChenM0C23} as the evidence is too large to naively ``read'', consume, and combine. Each claim is paired with a unique database---out of 80 in total---that corresponds to an evidence context of roughly 110M tokens. More specifically, each claim involves 11 tables and 4.5M records on average, reflecting the scale and complexity of real-world fact-checking.



\begin{figure*}[htbp]
  \includegraphics[width=\textwidth]{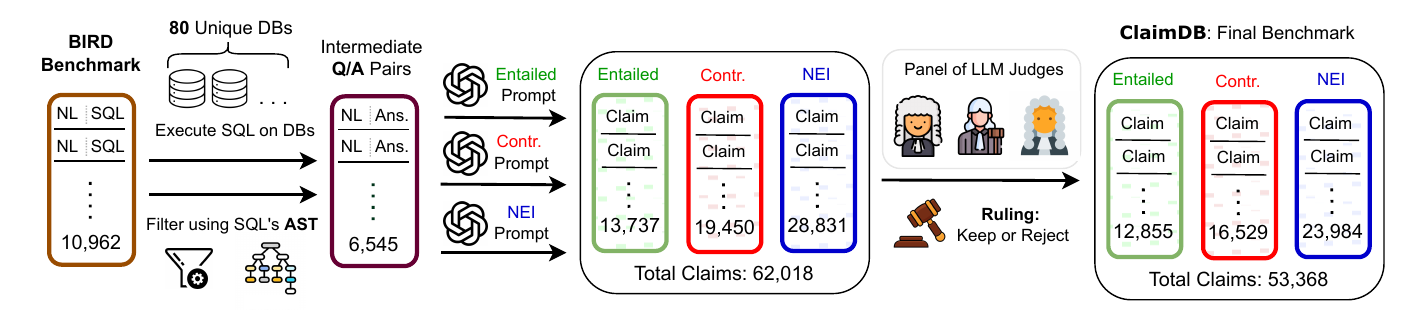}
  \caption{Overview of the \system construction pipeline. We start from the NL-to-SQL BIRD benchmark (Section~\ref{sec:bird}), execute each query on its respective real-world database, and filter out \textit{low-information} pairs using the query AST (Sections~\ref{sec:filtering}, \ref{sec:qapairs}). For each remaining Q/A pair, we prompt \texttt{gpt-5} to generate claims grounded in the gold answer---with some additional context for NEI claims (Section~\ref{sec:claim_gen}). We then use a panel of LLM judges from Mistral AI, Microsoft, and xAI to retain only high-quality claims (Section~\ref{sec:judges}). Finally, we apply embedding-based post-processing to two NEI categories for high-quality sampling (Section~\ref{sec:embeddings}); this step is omitted for space reasons.}
  \label{fig:workflow_claimdb}
\end{figure*}


The remainder of the paper is organized as follows. Section~\ref{sec:related} reviews existing fact-checking benchmarks. Sections~\ref{sec:benchmark}--\ref{sec:splits} describe the full benchmark construction pipeline, including claim generation, quality control with LLM judges, and dataset splits. We then evaluate 30 LLMs under a specific neuro-symbolic architecture in Section~\ref{sec:eval}, before concluding the paper in Section~\ref{sec:conclusion}.






\section{Related Work}
\label{sec:related}

Fact verification started with free-form text. The seminal work of \citet{DBLP:conf/naacl/ThorneVCM18} on FEVER pairs claims with Wikipedia sentences to determine whether they are supported or refuted. LIAR~\cite{wangLiarLiarPants2017} is a complementary benchmark of political claims with fine-grained veracity labels. Since then, fact-verification has evolved beyond text to include structured and semi-structured evidence. FEVEROUS~\cite{DBLP:conf/nips/AlyGST00CM21} extends FEVER with tables, while \citet{DBLP:journals/pacmmod/BussottiVSP23, DBLP:conf/emnlp/BussottiRFMP24} automate training-data generation for such settings. Other works explore verification over time-series data~\cite{strong-vlachos-2025-tsver}, Wikipedia info-boxes~\cite{DBLP:conf/acl/GuptaMNS20}, knowledge graphs~\cite{DBLP:conf/emnlp/DammuNDKRCS24}, financial reports~\cite{DBLP:conf/emnlp/0001LJWCLTZZC24}, temporal data~\cite{DBLP:conf/emnlp/BarikHL24, DBLP:conf/ijcai/BarikHL25}, scientific paper abstracts~\cite{DBLP:conf/emnlp/WaddenLLWZCH20}, U.S. court rulings~\cite{DBLP:journals/corr/abs-2601-17230}, and combined text and visual representations of tabular data (e.g., charts)~\cite{DBLP:conf/acl/ZhouZHCHC25}.

As the evidence modality naturally evolves to \textit{real-life}, large structured sources like tables, the reasoning starts to become more symbolic. The popular TabFact benchmark~\cite{DBLP:conf/iclr/ChenWCZWLZW20} requires operations like aggregation (e.g., \texttt{AVG}, \texttt{SUM}, \texttt{MAX}) and comparison. Subsequent benchmarks, including SEM-TAB-FACTS~\cite{DBLP:conf/semeval/WangMDR21}, and \mbox{SCITAB}~\cite{DBLP:conf/emnlp/LuPLNK23} further explored this setting by grounding real-life claims in tables extracted from complex scientific articles. 

However, existing fact-verification benchmarks rely on tables that are small enough to fit within an LLM's context window. Evidence is drawn almost exclusively from Wikipedia or scientific articles, where tables are inherently limited in size. Recent work by \citet{DBLP:conf/acl/DevasierPM025, DBLP:journals/corr/abs-2601-17232} begins to explore larger structured data, but these approaches remain limited---either to small-scale \emph{pilot} claim sets or to single-table evidence. In \system, we go further by designing a benchmark where claims are grounded in databases with millions of records, requiring multi-hop and compositional reasoning across tables. At this scale and complexity, verification can no longer rely on naively ``reading'' and then reasoning over in-context evidence. 



\section{\system Benchmark}
\label{sec:benchmark}
In \system, the task is to determine whether a given claim is \circled{1} \textit{\textcolor{e}{entailed}}, \circled{2} \textit{\textcolor{c}{contradicted}}, or \circled{3}~has \textit{\textcolor{nei}{not-enough-info}} (\textcolor{nei}{NEI}) with respect to the evidence in a database. Each claim is paired with exactly one large, multi-table database, which serves as the sole source of evidence for verification.

The workflow for creating the \system benchmark is shown in Figure~\ref{fig:workflow_claimdb}. In this section, we focus on the first stage of this process---the creation of claims---followed by quality-control in Section~\ref{sec:quality:control}.


\begin{table}[t]
\centering
\small
\begin{tabular}{l c}
\toprule
\textbf{Property} & \textbf{Value} \\
\midrule
Total Claims & 53,368 \\
\midrule
\multicolumn{2}{l}{\textbf{\textit{Label distribution}}} \\
\addlinespace[0.1em]
\# \textcolor{e}{Entailed} (\textcolor{e}{E}) & 12,855 \\
\# \textcolor{c}{Contradicted} (\textcolor{c}{C}) & 16,529 \\
\# \textcolor{nei}{Not Enough Info} (\textcolor{nei}{NEI}) & 23,984 \\
\addlinespace[0.15em]
\multicolumn{2}{l}{\quad \textit{NEI subcategories}} \\
\addlinespace[0.1em]
\quad \# Out-of-Schema & 12,644 \\
\quad \# Counterfactual & 5,786 \\
\quad \# Subjective & 5,554 \\
\midrule
\multicolumn{2}{l}{\textbf{\textit{Claim Context (Evidence)}}} \\
\addlinespace[0.1em]
Databases per Claim & 1 \\
Avg. Tables per Claim & \textbf{11.3} \\
Avg. Records per Claim & \textbf{4.6M} \\
Avg. Tokens per Claim$^{\dagger}$ & \textbf{110M} \\
\midrule
\addlinespace[0.3em]
Total Databases (DBs) & 80 \\
\bottomrule
\end{tabular}
\caption{Summary statistics of \system. $^{\dagger}$Token counts per claim are computed using the tokenizer of the \texttt{gpt-5} series as reference (see Appendix~\ref{appendix:db-to-tokens} for details).}
\label{tab:benchmark_stats}
\end{table}

\begin{figure}[t]
\centering
\includegraphics[width=0.9\columnwidth]{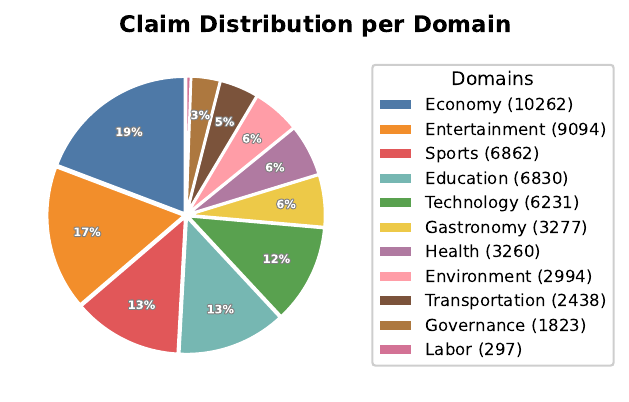}
\scriptsize
\begin{tabular}{ll}
\toprule
\textbf{Domain} & \textbf{Subdomains} \\
\midrule
Economy & Finance, World Economies, Retail, Banking \\
Entertainment & Movies, Music, TV Shows, Games, Cartoons \\
Sports & Basketball, Olympics, Hockey, F1, Soccer \\
Technology & Software, IT, Blockchain, Vision \\
Education & University, Academia, Schools, Language, Books \\
Gastronomy & Food, Restaurant \\
Health & Healthcare, Medical, Biology, Chemistry \\
Environment & Weather, Geography \\
Transportation & Transit Systems, Airport \\
Governance & Crime, Law \\
Labor & Human Resources \\
\bottomrule
\end{tabular}

\caption{Claim distribution and taxonomy. We group the 80 databases into 11 high-level domains (introduced in this work), each comprising multiple subdomains. The subdomains are inherited from \citet{DBLP:conf/nips/LiHQYLLWQGHZ0LC23} with a few minor modifications.}
\label{fig:domain_distribution_plot}
\end{figure}

\subsection{Overview}

The final \system benchmark contains 53,368 claims paired with databases from the BIRD benchmark (Section~\ref{sec:bird}): 12,855 \textit{\textcolor{e}{entailed}}, 16,529 \textit{\textcolor{c}{contradicted}}, and 23,984 \textit{\textcolor{nei}{not-enough-info}}. Table~\ref{tab:benchmark_stats} reports a detailed breakdown of label distributions and dataset statistics.



What makes \system challenging is the scale of the underlying data. Each claim is paired with a database containing, on average, 11.3 tables and 4.6M records. When converted to text, this amounts to roughly 110M tokens of evidence per claim (Appendix~\ref{appendix:db-to-tokens}). This exceeds the context window of modern LLMs by two to three orders of magnitude---for reference, Google's flagship \emph{long-context} models support up to 1M--2M tokens~\cite{Google-2025-Gemini-3-flash}. Thus, common approaches that rely on feeding tables directly into an LLM (e.g., \citealp{DBLP:conf/iclr/0002ZLEP0MFSLP24}) are no longer viable. 

Instead, successful systems must rely on \textit{executable} programs to carry out the heavy compositional reasoning required by \system over large volumes of data---for example, averaging or sorting millions of records (Section~\ref{sec:filtering}). Promising directions include Program of Thoughts (PoT)~\cite{DBLP:journals/tmlr/ChenM0C23}, tool-calling agents~\cite{DBLP:journals/tmlr/WangX0MXZFA24, theologitisThucyLLMbasedMultiAgent2025}, and Recursive Language Models~\cite{zhang2026recursivelanguagemodels}.

The benchmark uses 80 distinct real-life databases from BIRD~\cite{DBLP:conf/nips/LiHQYLLWQGHZ0LC23}, covering a wide range of domains and subdomains, from governance and healthcare to media, education, transportation, and the natural sciences. Figure~\ref{fig:domain_distribution_plot} summarizes the domain taxonomy present in \system (full details are in Appendix Table~\ref{tab:full_domain_taxonomy}).

\subsection{BIRD Benchmark}\label{sec:bird}

Our starting point is the BIRD~\cite{DBLP:conf/nips/LiHQYLLWQGHZ0LC23} NL2SQL benchmark.  Mapping NL to SQL has a long history, dating back to early work on NL interfaces~\cite{DBLP:journals/coling/WarrenP82}. More recent systems include WikiSQL~\cite{DBLP:journals/corr/abs-1709-00103}, Spider~\cite{DBLP:conf/emnlp/YuZYYWLMLYRZR18}, and KaggleDBQA~\cite{DBLP:conf/acl/LeePR20}.  BIRD is the first benchmark that consists of realistic databases, with multiple, large tables (millions of rows), including ``dirty'' data such as null values and inconsistent formatting. 

BIRD provides us with 11k NL-to-SQL examples (the combined public \texttt{train} and \texttt{test} splits). Each example is paired with the underlying database and accompanied by \textit{external information} specific to each example. For example, in the \texttt{toxicology} database, a label marked with ``$+$'' indicates that a molecule is carcinogenic, while ``$-$'' indicates the opposite. Throughout all subsequent transformations, we preserve any \textit{external info} intact. Importantly for us, the NL-to-SQL pairs have been carefully curated and vetted by trained, independent annotators, and the SQL queries are produced via a double-blind annotation process with experts in the loop~\cite{DBLP:conf/nips/LiHQYLLWQGHZ0LC23}.

\subsection{Pre-Filtering}\label{sec:filtering}

We start from the 11k NL-to-SQL pairs provided by BIRD. As a first filtering step, we retain only pairs where the SQL query combines substantial information in order to answer the NL question. For example, a query that involves a superlative (e.g., \texttt{MAX}) or averaging (\texttt{AVG}) summarizes large portions of the database, collapsing millions of values into one. This is highly desirable for us, because verifying claims derived from such pairs implicitly requires compositional reasoning over the database. On the other hand, a lookup query that returns a single record (e.g., \textit{``What is the name of the mayor of Seattle?''}) is of no interest: a verifier could answer by simply inspecting and seeing the \emph{right} data.



\vspace{1mm}
\noindent \textbf{Compositional Queries.} To identify such cases, we convert every SQL query into an abstract syntax tree (AST) and use it to identify the ones that contain computations over substantial parts of the data. More specifically, a query is retained if and only if its AST contains \emph{at least one} of the following:
\begin{itemize}[label=$\circ$, leftmargin=1em, rightmargin=1em, itemsep=0pt, topsep=2pt, parsep=0pt]
    \item Orderings or superlatives, which introduce comparisons over all data (e.g., \texttt{ORDER BY}).
    \item Aggregate Functions, such as \texttt{AVG} and \texttt{SUM}, which operate over large sets of records.
    \item Window functions, which involve complex information flow across partitions of the data.
    \item Multi-table joins, where three or more tables are combined for the final result.
\end{itemize}
The common theme is that answering the question (or, later, verifying the claim) requires combining information from large portions of the database, which will far exceed the context size of any LLM. Figures~\ref{fig:q7}--\ref{fig:q9} in  Appendix~\ref{sec:appendix_qast} provide concrete examples of queries, their ASTs, and the corresponding filtering decisions following the rules above.

Finally, we discard queries whose answers return more than ten records. These answers are later fed to an LLM during claim generation (Section~\ref{sec:claim_gen}), and large result sets would make it harder for the model to track the structure (columns and values). This ten-record cutoff is conservative for modern models like \texttt{gpt-5}, but helps ensure that errors do not arise from misinterpreting the structured data.


\vspace{1mm}
After these two filtering steps we end up with roughly 6.5k NL/SQL pairs, where each SQL query ``touches'' a large subset of its associated database.

\subsection{Question-Answer Pairs over Databases}\label{sec:qapairs}

Next, we execute each SQL query on its corresponding database to obtain the exact answer. This yields Question/Answer pairs grounded in real-world databases. From this point on, the SQL itself is no longer needed and can be discarded. 

The answer to each SQL query is a table, i.e. a set of rows, which is not a friendly format for LLMs.  Following~\citet{singha2023tabular}, we convert each answer from a table to JSON, a format that is friendlier for LLMs. After this step we are left with 6.5k Question/JSON-Answer pairs.

\subsection{Claim Generation}\label{sec:claim_gen}
Each pair serves as ground truth. Given one such pair, we prompt \texttt{gpt-5} to generate claims that are either \textit{\textcolor{e}{entailed}} by the answer or \textit{\textcolor{c}{contradicted}} by it (we discuss \textcolor{nei}{NEI} claims later in this section). Figure~\ref{fig:claim_example} shows a concrete example of a single Q/A pair together with the claims generated from it.

\begin{figure}[htbp]
\centering
\fbox{
\begin{minipage}{0.95\columnwidth}
\small
\textbf{Domain:} California Schools \\
\textbf{Question:} Which cities have the top 5 lowest enrollment number for students in grades 1 through 12? \\
\textbf{Answer:} Coulterville; Pinecrest; Shaver Lake; Emigrant Gap; Hyampom. \\[-0.45em]

\textbf{Generated Claims:}
\begin{itemize}[leftmargin=1em, itemsep=0.0em, parsep=2pt, topsep=2pt]
    \item \textbf{\textcolor{e}{Entailed}:} In California, the five cities with the lowest \mbox{K-12} student enrollment are Coulterville, Pinecrest, Shaver Lake, Emigrant Gap, and Hyampom.
    \item \textbf{\textcolor{c}{Contradicted}:} The three California cities with the lowest K-12 enrollments are Lee Vining, Trona and Keeler.
    \item \textbf{\textcolor{nei}{NEI} (Out-of-Schema):} Most families in Hyampom choose homeschooling rather than enrolling their children in public schools.
    \item \textbf{\textcolor{nei}{NEI} (Counterfactual):} If Pinecrest had opened a new K-12 campus in 2020, its grades 1-12 enrollment would have increased enough that the city would not rank among the five lowest in California.
    \item \textbf{\textcolor{nei}{NEI} (Subjective):} Among Coulterville, Pinecrest, Emigrant Gap, and Hyampom, Emigrant Gap provides the most \textit{nurturing} learning environment for K--12 students.
\end{itemize}
\end{minipage}
}
\caption{Example of claims generated from a single Q/A pair in \system (California Schools database). In the generation of \textcolor{nei}{NEI} claims we also give the database \textit{schema} information along the golden context.}
\label{fig:claim_example}
\end{figure}

For this step we explored a range of prompting strategies~\cite{Kotha-2025-OpenAI-Prompting-Guide}, including different task instructions, zero-shot versus few-shot, and varying levels of CoT, while keeping track of failure modes---which we deal with in the next section.

Our final setup uses a single prompt for each label (3 in total). All final prompts are available in Figures~\ref{fig:contradicted_prompt}, \ref{fig:entailed_prompt} and \ref{fig:nei_prompt} of the Appendix. 

For both \textit{\textcolor{e}{entailed}} and \textit{\textcolor{c}{contradicted}} claims, we found that 1-shot prompting with medium reasoning works best. Given a Q/A pair, the model is instructed to generate between one and three claims.

\vspace{1mm}
\noindent \textbf{NEI Claims.} Lastly, we generate claims that are \textit{definitely unanswerable} from the database at hand. We draw inspiration from the taxonomy presented by \citet{DBLP:journals/corr/abs-2506-09038} and the work of \citet{DBLP:conf/acl/AmayuelasWPCW24}. Given our specific setting, we can safely support three claim categories.

First, we generate claims that fall outside the concepts of the database. To do this, we inspect the database schema metadata offline---including table and column names, and relationships---and provide the full schema information to \texttt{gpt-5}. We then explicitly instruct the model (Figure~\ref{fig:nei_prompt} in the Appendix) to produce claims that are based on concepts not represented anywhere in the schema. The model has no trouble doing this effectively: for example, in Figure~\ref{fig:claim_example}, the generated \textit{out-of-schema} claim refers to \textit{homeschooling}---a concept that is not present in the California Schools database.


Second, we generate \textit{subjective} claims which express opinions or value judgments that cannot be objectively verified. Finally, we generate \textit{counterfactual} claims. These describe imagined or ``what if'' scenarios that are unanswerable with certainty.

For \textcolor{nei}{NEI} claim generation, we found that zero-shot prompting with medium reasoning works best. There is a large spectrum of possible and appropriate \textcolor{nei}{NEI} claims, as they are largely unconstrained by the concepts of the data. As a result, any form of in-context learning biases the model toward specific patterns, limiting generation diversity overall. In contrast, \texttt{gpt-5} naturally generates a diverse set of \textcolor{nei}{NEI} claims in the zero-shot setting.


\vspace{2mm}
All in all, we end up with roughly \textcolor{e}{14k}, \textcolor{c}{19k}, and \textcolor{nei}{28k} \textcolor{e}{entailed}, \textcolor{c}{contradicted}, and \textcolor{nei}{NEI} claims, respectively (Table~\ref{tab:benchmark_stats}). The discrepancy in counts arises because \textcolor{c}{contradicted} claims are not constrained by the answer itself (they can oppose it arbitrarily), while \textcolor{nei}{NEI} claims are not constrained by either the answer or even the concepts present in the Q/A pair or the database (we address this in Section~\ref{sec:embeddings}).

\section{Quality Control}
\label{sec:quality:control}

Automatic claim generation is never perfect. Issues can still slip-in such as claims that reference prior context---which is opaque to a verifier---or leak helpful information. While outright label errors are rare (the claim generation task is trivial), we take \textit{conservative} steps to filter problematic claims.
This is the second half of the workflow in Figure~\ref{fig:workflow_claimdb}.

\subsection{Panel of LLM Judges}\label{sec:judges}

Manually annotating 64k examples is infeasible, so we rely on the now widely adopted \textit{LLM-as-a-Judge} paradigm, where an LLM evaluator is used as a proxy for human annotators.

\vspace{1.5mm}
\noindent \textbf{Judge Panel.} It is well-known that LLMs tend to prefer answers generated by themselves~\cite{DBLP:conf/nips/ZhengC00WZL0LXZ23, DBLP:conf/iclr/YeWHCZMGG0CC025}. To avoid this self-enhancement bias, we therefore exclude OpenAI models entirely. We further follow \citet{DBLP:journals/corr/abs-2404-18796}, who show that replacing a single large evaluator with a \textit{panel of judges} of smaller LLMs leads to more stable results. So, we construct a judge panel from three different model families coming from Microsoft, xAI, and Mistral AI:
\begin{enumerate}[label=(\alph*), leftmargin=3em, itemsep=0pt, topsep=2pt, parsep=0pt]
    \item \texttt{Phi-4}~\cite{DBLP:journals/corr/abs-2412-08905}
    \item \texttt{grok-3-mini}~\cite{xAI-2025-grok-3}
    \item \texttt{mistral-small}~\cite{MistralAI-2025-Small} \eat{DBLP:journals/corr/abs-2310-06825}
\end{enumerate}

\noindent Each judge independently evaluates every claim, and we later combine their judgements for the final \textit{ruling}---whether a claim is eliminated or not. Importantly, we also ask the judges to \textit{justify} their reasoning and verdicts in a few sentences in NL.

\vspace{1.5mm}
\noindent \textbf{Rubrics.} Before putting the judge panel to work, we must first clearly define the evaluation criteria. Each judge is given the full gold context---namely the NL question, answer, and domain---along with the generated claim and its assigned label. Judges then answer the following binary (``yes'' or ``no'') questions: \circled{1} \textit{Is the label correct?}, and \circled{2} \textit{Is the claim self-contained?} The second question refers to claims where there are opaque references to previous context (e.g., ``the question above'', etc.).

\textcolor{nei}{NEI} claims require additional care. During generation, \texttt{gpt-5} had access to the schema as part of the gold context. As a result, schema artifacts occasionally appear directly in the generated claims, which unintentionally helps systems evaluated on the benchmark. For these claims, we therefore ask two additional questions: \circled{3} \textit{Is there schema leakage?}, and \circled{4} \textit{Is the assigned \textcolor{nei}{NEI} category correct?}

\vspace{1mm}
\noindent \textbf{Ruling.} If \emph{any} judge flags a claim as problematic under \emph{any} rubric (e.g., incorrect label, lack of self-containment, etc.), the claim is discarded. In other words, a single negative judgment is grounds for elimination. This is intentionally \emph{conservative} as it gives us confidence that the final benchmark is clean---even if that means over-eliminating claims.

\vspace{1mm}
\noindent \textbf{Prompt.} The evaluator prompt plays a central role in how reliable the judging process is~\cite{DBLP:conf/iclr/YeWHCZMGG0CC025}. We already simplify the task as much as possible (e.g., binary classification rubrics), and we make one more simplifying decision: we are willing to ``unfairly'' discard borderline claims.

While most LLM-as-a-Judge setups aim to maximize agreement with human annotators~\cite{DBLP:journals/corr/abs-2406-12624}, our goal is slightly different. We want to maximize the \textit{recall} on claims that humans would eliminate. To that end, the prompt includes instructions such as \textit{``If you are unsure, answer no.''}

To calibrate the prompt, we manually annotated 300 examples (150 \textcolor{nei}{NEI} and 150 \textcolor{c}{contradicted} or \textcolor{e}{entailed}) using our rubrics. We shuffle and split these annotated examples in two halves: one for calibrating the prompt, and the other for testing it.

\vspace{1mm}
\noindent \textbf{Prompt Engineering.} After calibration, we end up with the prompts shown in Figures~\ref{fig:con_entailed_judge_prompt} and~\ref{fig:nei_judge_prompt} of the Appendix. The former is used for \textcolor{e}{entailed} and \textcolor{c}{contradicted} claims, while the latter for \textcolor{nei}{NEI} claims and includes the two additional evaluation rubrics.

\vspace{1mm}
\noindent \textbf{Prompt Results.} We evaluate the judge panel on the human-annotated test set (75 \textcolor{nei}{NEI} and 75 \textcolor{c}{contradicted} or \textcolor{e}{entailed}), which contains 4 claims identified as problematic by PhD annotators. The panel successfully discards all four, achieving $100\%$ recall, and behaves conservatively. We also did not observe any mislabeled claims in terms of veracity in this set. This aligns with our experience during the construction of the benchmark---generating claims from correct Q/A pairs is an easy task, and \texttt{gpt-5} is more than capable of doing so reliably.



\vspace{2mm}
Finally, we run the full benchmark through the judge panel with all three models evaluated at temperature zero. Overall, $5.6\%$ of claims are flagged for incorrect labels and $1.7\%$ for not being self-contained. Among \textcolor{nei}{NEI} claims, $11\%$ show schema leakage and $5\%$ have an incorrect category assignment. Taken together, this filtering step removes approximately $14\%$ of the benchmark (Figure~\ref{fig:workflow_claimdb}).

\subsection{Grounding NEI Claims}\label{sec:embeddings}

The main reason we have a large number of \textcolor{nei}{NEI} claims is that they are inherently unconstrained by the database concepts. They can be about almost anything, especially in the case of \textit{out-of-schema} claims. However, this freedom often makes them very obvious to classify. The same issue arises with \textit{counterfactual} claims---given a bit of imagination (\texttt{gpt-5} does not lack this quality), it is easy to go overboard. These overboard claims are not bad per se; we simply want a way to avoid sampling them when we construct our main benchmarking test set. In this section, we create a way to filter out claims based on how \textit{far} they are from the underlying database concepts. We later use this signal to construct a more difficult test set (Section~\ref{sec:splits}).

Essentially, we are looking for a way to rank NL claims by how ``close'' they are to the concepts of the underlying data---we can approximate this using the golden context. We capture this ``closeness'' using semantic textual similarity (STS). More specifically, we embed\footnote{We use \texttt{gemini-embedding-001}~\cite{DBLP:journals/corr/abs-2503-07891} for the embeddings; at this time, it ranks 4\textsuperscript{th} on MTEB~\citeyearpar{DBLP:conf/eacl/MuennighoffTMR23}.} the Q/A pair and the generated claim, and compute the similarity between them.  The resulting score gives us a simple and effective way to \emph{rank} claims by how closely they stay grounded in plausible concepts of the database.

\begin{figure}[htbp]
    \centering
  \includegraphics[width=0.88\columnwidth]{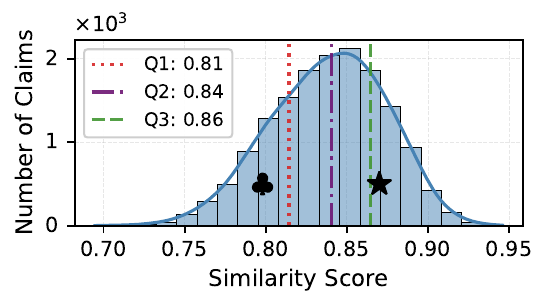}
  \caption{Distribution of similarity scores between generated claims and their gold context for \textit{out-of-schema} claims. Higher scores indicate that claims stay ``closer'' to the underlying data concepts. The two highlighted claim examples ($\clubsuit$, $\bigstar$) are discussed in Section~\ref{sec:embeddings}.}
  \label{fig:similarity_ofs}
\end{figure}

Figure~\ref{fig:similarity_ofs} shows the distribution of similarity scores for \textit{out-of-schema} claims. To make this concrete, we look at two claims generated from the same Q/A in the Chicago Crime database, which asks about crimes in Chicago's Central district. Although both claims are \emph{out-of-schema}, they fall into very different parts of the similarity distribution:
\begin{itemize}[label={}, leftmargin=2em, rightmargin=2em, topsep=1.5pt, itemsep=1pt, parsep=0pt]
\item ($\clubsuit$) \textit{The commander of Chicago’s Central police district holds a law degree from an Illinois university.}
\item ($\bigstar$) \textit{Crimes recorded in Chicago’s Central district disproportionately involve tourists compared to other districts.}
\end{itemize}

The second claim is much closer to the database concepts: validating it would require inspecting the data to see whether victim information (e.g., registered residents) is available, which is at least plausible. The first claim, in contrast, drifts far from what the database could reasonably support. This difference is captured by the similarity score: the second claim ($\bigstar$) falls in the top quartile (0.869), while the first ($\clubsuit$) lies in the lowest quartile (0.798).

\section{Benchmark Splits}\label{sec:splits}
We divide \system into three disjoint splits: \circled{1}~the training set, \circled{2} a \emph{public} test set, and \circled{3}~a \textit{private} test set (the labels are hidden). Both test sets contain approximately 1,000 examples each. They are sampled\footnote{The test sets are drawn from claims coming from the \texttt{dev} split of BIRD as it has received substantial cleanup effort in prior work (e.g.,~\citealp{DBLP:conf/acl/WretbladRBAH24, DBLP:conf/kdd/LiuSL0L25}).} uniformly at random per label and fully balanced, with roughly 333 examples per label. For \textcolor{nei}{NEI} claims, the three subcategories are also balanced with roughly 111 examples each; for both \textit{out-of-schema} and \textit{counterfactual}, sampling is further restricted to the top quartile of the similarity score distribution (Figures~\ref{fig:similarity_ofs} and \ref{fig:similarity_appendix}). In contrast, the training split contains the full spectrum of \textcolor{nei}{NEI} claims. We release similarity scores for all \textcolor{nei}{NEI} claims, enabling the construction of any custom downstream training split---for example, one that mirrors the test set distribution.

\section{Evaluation}\label{sec:eval}

\subsection{Setup}\label{sec:setup}
Claims in \system are intentionally derived from insights that \emph{combine} information across millions of records through operations such as averaging, grouping and sorting (Section~\ref{sec:filtering}). At this scale, verification cannot be done by ``reading'' the evidence in-context: the claim context contains, on average, 110M tokens of structured information---which is far beyond any LLM's context window. As a result, verifiers must reason in \textit{executable} programs that handle, symbolically, the bulk of the compositional reasoning that is required.

Our evaluation primarily focuses on tool-calling agents~\cite{DBLP:conf/iclr/MialonF0LS24, DBLP:journals/tmlr/WangX0MXZFA24}, which we view as a very strong and practical solution for this setting. We use Google's MCP toolbox for databases~\cite{Google-2025-ToolBox} and give each agent a single tool: the ability to execute arbitrary SQL queries against the database associated with the claim at hand. The only constraint we impose is a \emph{generous} limit of 20 tool calls; exceeding this counts as a failed run.

There also exist other setups such as coding agents~\cite{DBLP:conf/icml/WangCY0L0J24} that interact with data via Python and pandas. However, prior work has shown such approaches to be brittle for the smaller LLMs (below 70B) we consider in this paper~\cite{DBLP:conf/coling/SunLL0ZLJ24}. We therefore leave a systematic study of potential alternatives---from other \emph{tool} choices to different paradigms altogether---for future work.

We use a single prompt (Figure~\ref{fig:verifier_prompt} in the Appendix) and we conduct extensive experiments with 30 state-of-the-art proprietary and open-source LLMs. Information and hyperparameters (e.g., decoding) are illustrated in Table~\ref{tab:hyperparams} of the Appendix.

\subsection{Metrics}

\system is a three-way classification benchmark with labels \textit{\textcolor{e}{entailed}} (\textcolor{e}{E}), \textit{\textcolor{c}{contradicted}} (\textcolor{c}{C}), and \textit{\textcolor{nei}{not-enough-info}} (\textcolor{nei}{NEI}). For each label, we compute precision, recall, and F1 by treating that label as the positive class and the remaining two as negatives.

\emph{Precision} measures the proportion of predictions for a given label that are correct, while \emph{recall} measures the fraction of true instances of that label that are correctly identified. F1 is their harmonic mean. We report all metrics, but we primarily focus on accuracy and $\text{Macro-F1} = \frac{1}{3}(\text{F1}_\text{\textcolor{e}{E}} + \text{F1}_\text{\textcolor{c}{C}} + \text{F1}_\text{\textcolor{nei}{NEI}})$.

\subsection{Discussion}


\begin{table*}[t]
\centering
\small
\setlength{\tabcolsep}{3.4pt}
\begin{tabular}{l@{\hspace{3pt}}
  c@{\hspace{1.5pt}}c@{\hspace{1.5pt}}c
  c@{\hspace{1.5pt}}c@{\hspace{1.5pt}}c
  c@{\hspace{6pt}}c@{\hspace{6pt}}c
  c c}
\toprule
\multirow{2}{*}{\textbf{LLM Backbone}\rule{0pt}{2.4ex}} &
\multicolumn{3}{c}{\textbf{Precision}} &
\multicolumn{3}{c}{\textbf{Recall}} &
\multicolumn{3}{c}{\textbf{F1}} &
\multirow{2}{*}{\textbf{Macro-F1}\rule{0pt}{2.6ex}} &
\multirow{2}{*}{\textbf{Acc.}\rule{0pt}{2.6ex}} \\
\cmidrule(lr){2-4}
\cmidrule(lr){5-7}
\cmidrule(lr){8-10}
& \textcolor{e}{Ent.} & \textcolor{c}{Contr.} & \textcolor{nei}{NEI}
& \textcolor{e}{Ent.} & \textcolor{c}{Contr.} & \textcolor{nei}{NEI}
& \textcolor{e}{Ent.} & \textcolor{c}{Contr.} & \textcolor{nei}{NEI}
& & \\

\midrule

\texttt{gpt-4o-mini} \citeyearpar{DBLP:journals/corr/abs-2410-21276}
& \gray{ 0.777 } & \gray{ 0.666 } & \gray{ 0.679 }
& \gray{ 0.577 } & \gray{ 0.656 } & \gray{ 0.107 }
& 0.662 & 0.661 & 0.185
& 0.503 & 0.445  \\

\texttt{gpt-4.1-nano} \citeyearpar{OpenAI-2025-gpt-4.1}
& \gray{ 0.647 } & \gray{ 0.641 } & \gray{ 0.376 }
& \gray{ 0.165 } & \gray{ 0.124 } & \gray{ 0.952 }
& 0.263 & 0.208 & 0.539
& 0.337 & 0.416 \\

\texttt{gpt-5-nano} \citeyearpar{OpenAI-2025-gpt-5}
& \gray{ 0.815 } & \gray{ 0.710 } & \gray{ 0.874 }
& \gray{ 0.742 } & \gray{ 0.900 } & \gray{ 0.720 }
& 0.777 & 0.794 & 0.790
& 0.787 & 0.787 \\

\texttt{gpt-5-mini} \citeyearpar{OpenAI-2025-gpt-5}
& \gray{ 0.875 } & \gray{ 0.713 } & \gray{ 0.963 }
& \gray{ 0.754 } & \gray{ 0.952 } & \gray{ 0.777 }
& 0.810 & 0.815 & \textbf{0.860}
& \textbf{0.828} & \textbf{0.827} \\

\texttt{gpt-oss:20b} \citeyearpar{openai2025gptoss120bgptoss20bmodel}
& \gray{ 0.851 } & \gray{ 0.735 } & \gray{ 0.676 }
& \gray{ 0.669 } & \gray{ 0.688 } & \gray{ 0.862 }
& 0.749 & 0.710 & 0.758
& 0.739 & 0.740 \\

\midrule

\texttt{gemini-2.5-flash} \citeyearpar{DBLP:journals/corr/abs-2507-06261}
& \gray{ 0.841 } & \gray{ 0.734 } & \gray{ 0.830 }
& \gray{ 0.685 } & \gray{ 0.825 } & \gray{ 0.869 }
& 0.755 & 0.777 & 0.849
& 0.793 & 0.793 \\

\texttt{gemini-3-flash} \citeyearpar{Google-2025-Gemini-3-flash}
& \gray{ 0.754 } & \gray{ 0.750 } & \gray{ 0.962 }
& \gray{ 0.799 } & \gray{ 0.934 } & \gray{ 0.673 }
& 0.776 & \textbf{0.832} & 0.792
& 0.800 & 0.801 \\

\midrule

\texttt{claude-3-haiku}  \citeyearpar{Anthropic-2024-Claude-3}
& \gray{ 0.572 } & \gray{ 0.556 } & \gray{ 0.433 }
& \gray{ 0.261 } & \gray{ 0.435 } & \gray{ 0.756 }
& 0.359 & 0.488 & 0.550
& 0.466 & 0.485 \\

\texttt{claude-3-5-haiku} \citeyearpar{Anthropic-2024-Claude-3}
& \gray{ 0.775 } & \gray{ 0.626 } & \gray{ 0.661 }
& \gray{ 0.580 } & \gray{ 0.743 } & \gray{ 0.702 }
& 0.663 & 0.680 & 0.681
& 0.675 & 0.675 \\

\texttt{claude-haiku-4-5} \citeyearpar{Anthropic-2025-Claude-Haiku-4.5}
& \gray{ 0.836 } & \gray{ 0.711 } & \gray{ 0.983 }
& \gray{ 0.796 } & \gray{ 0.952 } & \gray{ 0.682 }
& \textbf{0.815} & 0.814 & 0.805
& 0.811 & 0.809 \\

\midrule

\texttt{qwen3:1.7b} \citeyearpar{DBLP:journals/corr/abs-2505-09388}
& \gray{ 0.645 } & \gray{ 0.533 } & \gray{ 0.351 }
& \gray{ 0.060 } & \gray{ 0.048 } & \gray{ 0.982 }
& 0.110 & 0.089 & 0.518
& 0.239 & 0.366 \\

\texttt{qwen3:4b} \citeyearpar{DBLP:journals/corr/abs-2505-09388}
& \gray{ 0.530 } & \gray{ 0.554 } & \gray{ 0.492 }
& \gray{ 0.450 } & \gray{ 0.218 } & \gray{ 0.860 }
& 0.487 & 0.312 & 0.626
& 0.475 & 0.511 \\

\texttt{qwen3:8b} \citeyearpar{DBLP:journals/corr/abs-2505-09388}
& \gray{ 0.785 } & \gray{ 0.617 } & \gray{ 0.450 }
& \gray{ 0.306 } & \gray{ 0.311 } & \gray{ 0.940 }
& 0.441 & 0.414 & 0.608
& 0.488 & 0.521 \\

\texttt{qwen3:14b} \citeyearpar{DBLP:journals/corr/abs-2505-09388}
& \gray{ 0.812 } & \gray{ 0.744 } & \gray{ 0.438 }
& \gray{ 0.285 } & \gray{ 0.360 } & \gray{ 0.943 }
& 0.422 & 0.485 & 0.599
& 0.502 & 0.531 \\

\texttt{qwen3:32b} \citeyearpar{DBLP:journals/corr/abs-2505-09388}
& \gray{ 0.732 } & \gray{ 0.667 } & \gray{ 0.491 }
& \gray{ 0.393 } & \gray{ 0.459 } & \gray{ 0.866 }
& 0.512 & 0.544 & 0.626
& 0.561 & 0.574 \\

\texttt{qwen3-coder:30b} \citeyearpar{DBLP:journals/corr/abs-2505-09388}
& \gray{ 0.744 } & \gray{ 0.625 } & \gray{ 0.660 }
& \gray{ 0.646 } & \gray{ 0.659 } & \gray{ 0.711 }
& 0.691 & 0.641 & 0.685
& 0.672 & 0.672 \\

\texttt{qwq:32b} \citeyearpar{QwenTeam-2025-QwQ}
& \gray{ 0.500 } & \gray{ 0.558 } & \gray{ 0.472 }
& \gray{ 0.336 } & \gray{ 0.508 } & \gray{ 0.667 }
& 0.402 & 0.532 & 0.552
& 0.495 & 0.504 \\

\midrule

\texttt{mistral-nemo:12b} \citeyearpar{MistralAI-2024-NeMo}
& \gray{ 0.380 } & \gray{ 0.446 } & \gray{ 0.381 }
& \gray{ 0.381 } & \gray{ 0.211 } & \gray{ 0.577 }
& 0.381 & 0.287 & 0.459
& 0.376 & 0.391 \\

\texttt{ministral-3:3b} \citeyearpar{MistralAI-2025-ministral-3}
& \gray{ 0.511 } & \gray{ 0.500 } & \gray{ 0.390 }
& \gray{ 0.276 } & \gray{ 0.163 } & \gray{ 0.821 }
& 0.359 & 0.246 & 0.529
& 0.378 & 0.422 \\

\texttt{ministral-3:8b} \citeyearpar{MistralAI-2025-ministral-3}
& \gray{ 0.633 } & \gray{ 0.557 } & \gray{ 0.484 }
& \gray{ 0.408 } & \gray{ 0.396 } & \gray{ 0.792 }
& 0.496 & 0.463 & 0.600
& 0.520 & 0.533 \\

\texttt{ministral-3:14b} \citeyearpar{MistralAI-2025-ministral-3}
& \gray{ 0.612 } & \gray{ 0.638 } & \gray{ 0.624 }
& \gray{ 0.598 } & \gray{ 0.580 } & \gray{ 0.690 }
& 0.605 & 0.608 & 0.655
& 0.623 & 0.623 \\

\texttt{mistral-small:22b} \citeyearpar{MistralAI-2025-Small}
& \gray{ 0.423 } & \gray{ 0.526 } & \gray{ 0.375 }
& \gray{ 0.222 } & \gray{ 0.060 } & \gray{ 0.878 }
& 0.291 & 0.108 & 0.525
& 0.308 & 0.389 \\

\texttt{magistral:24b} \citeyearpar{DBLP:journals/corr/abs-2506-10910}
& \gray{ 0.682 } & \gray{ 0.714 } & \gray{ 0.419 }
& \gray{ 0.309 } & \gray{ 0.257 } & \gray{ 0.905 }
& 0.426 & 0.378 & 0.573
& 0.459 & 0.492 \\

\texttt{devstral:24b} \citeyearpar{MistralAI-2025-Devstral}
& \gray{ 0.415 } & \gray{ 0.624 } & \gray{ 0.399 }
& \gray{ 0.291 } & \gray{ 0.221 } & \gray{ 0.771 }
& 0.342 & 0.326 & 0.526
& 0.398 & 0.429 \\

\texttt{devstral-small-2} \citeyearpar{rastogi2025devstralfinetuninglanguagemodels}
& \gray{ 0.602 } & \gray{ 0.712 } & \gray{ 0.521 }
& \gray{ 0.610 } & \gray{ 0.447 } & \gray{ 0.705 }
& 0.606 & 0.549 & 0.599
& 0.585 & 0.588 \\

\midrule

\texttt{nemotron-3-nano:30b} \citeyearpar{nvidia2025nvidianemotron3efficient}
& \gray{ 0.714 } & \gray{ 0.726 } & \gray{ 0.612 }
& \gray{ 0.652 } & \gray{ 0.601 } & \gray{ 0.747 }
& 0.681 & 0.658 & 0.673
& 0.671 & 0.667 \\

\midrule

\texttt{llama3.1:8b} \citeyearpar{grattafiori2024llama3herdmodels}
& \gray{ 0.341 } & \gray{ 0.433 } & \gray{ 0.337 }
& \gray{ 0.222 } & \gray{ 0.079 } & \gray{ 0.726 }
& 0.269 & 0.133 & 0.461
& 0.288 & 0.344 \\

\texttt{llama3.2:3b} \citeyearpar{grattafiori2024llama3herdmodels}
& \gray{ 0.428 } & \gray{ 0.412 } & \gray{ 0.366 }
& \gray{ 0.372 } & \gray{ 0.100 } & \gray{ 0.685 }
& 0.398 & 0.161 & 0.477
& 0.345 & 0.387 \\

\midrule

\texttt{cogito:14b} \citeyearpar{DeepCogito-2025-Cogito}
& \gray{ 0.758 } & \gray{ 0.701 } & \gray{ 0.384 }
& \gray{ 0.141 } & \gray{ 0.184 } & \gray{ 0.973 }
& 0.238 & 0.292 & 0.551
& 0.360 & 0.435 \\

\texttt{cogito:32b} \citeyearpar{DeepCogito-2025-Cogito}
& \gray{ 0.797 } & \gray{ 0.564 } & \gray{ 0.506 }
& \gray{ 0.354 } & \gray{ 0.556 } & \gray{ 0.792 }
& 0.491 & 0.560 & 0.617
& 0.556 & 0.568 \\

\bottomrule
\end{tabular}
\caption{We report per-label precision, recall, and F1, along with macro-F1 and accuracy.}
\label{tab:full-results}
\end{table*}

\begin{figure}[t]
  \includegraphics[width=0.95\columnwidth]{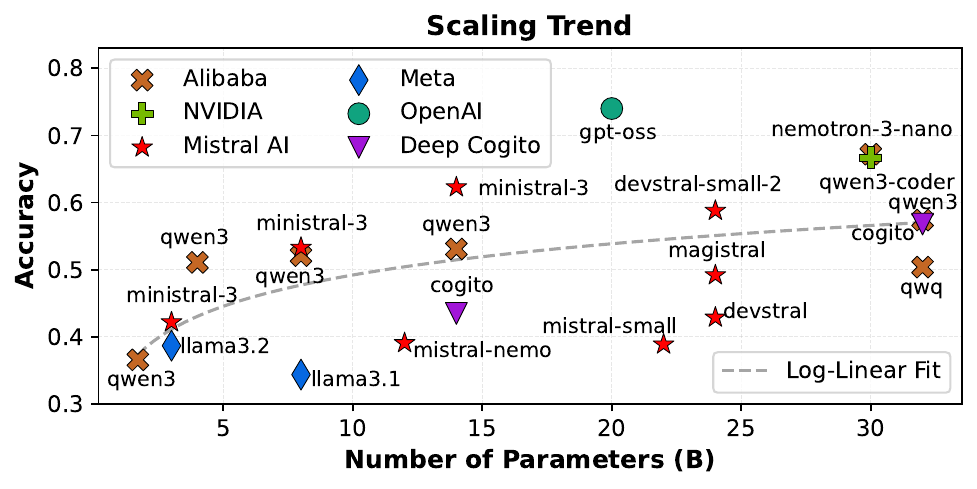}
  \caption{Final accuracy vs. model size (open-source).}
  \label{fig:scaling_trend}
\end{figure}

\noindent \textbf{Scaling Trends.} Figure~\ref{fig:scaling_trend} plots accuracy against model size. We notice the familiar pattern of larger models performing better, but with improvements that are marginal and roughly log-linear.

\begin{figure}[t]
  \includegraphics[width=0.95\linewidth]{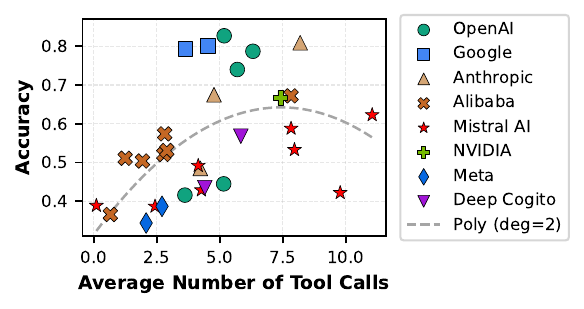}
  \caption{Average number of tool calls per model on the public \texttt{test} set of \system. Performance generally degrades as LLM-database interactions become longer.}
  \label{fig:average_number_of_tools_calls}
\end{figure}

\vspace{0.5mm}
\noindent \textbf{Long Sessions Hurt.} Long LLM-database interactions have inherent fragility: as they grow longer, models increasingly lose focus due to the large amounts of information. In this setting, a single bad decision---a single \textit{careless} query---can flood the model with hundreds of thousands of tokens.

Figure~\ref{fig:average_number_of_tools_calls} plots, for each of the 30 models, the \textit{average number of tool calls} over the 1,000 \texttt{test} examples. We fit a second-order polynomial and observe that the top-performing models typically average roughly 4--8 tool calls. Longer (or shorter) than that leads to degraded performance. The complete tool-call distributions are shown in Figure~\ref{fig:tool_calls_histogram}.


\vspace{0.5mm}
\noindent \textbf{Open-Source Models Struggle.} Table~\ref{tab:full-results} reports results for all 30 models. Across both accuracy and macro-F1, \texttt{gpt-5-mini} performs best, followed by \texttt{claude-haiku-4.5} and \texttt{gemini-3-flash}. Out of all evaluated models, more than half (17 out of 30) stay below 55\% accuracy and Macro-F1.

Open-source models generally struggle, with the exception of \texttt{gpt-oss-20b} which  performs on par with top proprietary models. Aside from \texttt{gpt-oss}, no other open-source model (out of the 20 remaining) exceeds 68\% accuracy or macro-F1. Overall, this highlights that there is room for improvement for open-source models---to catch up---in reasoning over large-scale data. The \system benchmark is a clear step toward this goal.

\begin{figure*}[htbp]
\centering
  \includegraphics[width=0.85\textwidth]{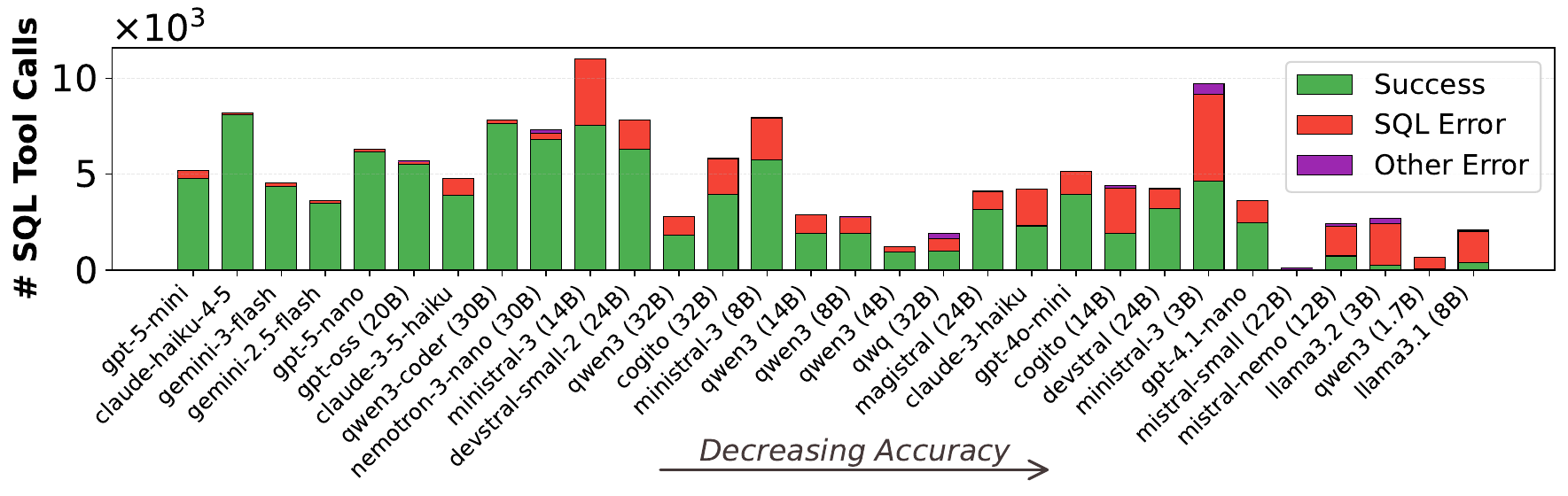}
  \caption{Our agents are equipped with a single tool: the ability to execute SQL queries on the associated database. We report the success rates of the SQL queries per model. More specifically, ``Success'' denotes queries that execute and return results; ``SQL Error'' denotes queries with syntax or logical issues; and ``Other Error'' includes all remaining failures. Notably, top-performing models achieve high query success rates, suggesting that failures are not superficial (e.g., syntax) but instead reflect deeper breakdowns in reasoning over large structured data.}
  \label{fig:tool_call_errors}
\end{figure*}

\vspace{0.5mm}
\noindent \textbf{No Evidence of Data Contamination.} A common concern with modern benchmarks is the risk of data contamination, where models may have seen parts of the data during training, making the benchmark obsolete~\cite{DBLP:conf/coling/SamuelZZ25}. To investigate this, we evaluate the best-performing model, \texttt{gpt-5-mini}, without access to external tools, where it must label each claim using only its parametric knowledge. We find that performance is near random, with a macro-F1 of 0.253 and accuracy of 0.367, suggesting that \system cannot be solved using parametric knowledge alone.

\begin{figure}[htbp]
  \centering
  \includegraphics[width=0.93\linewidth]{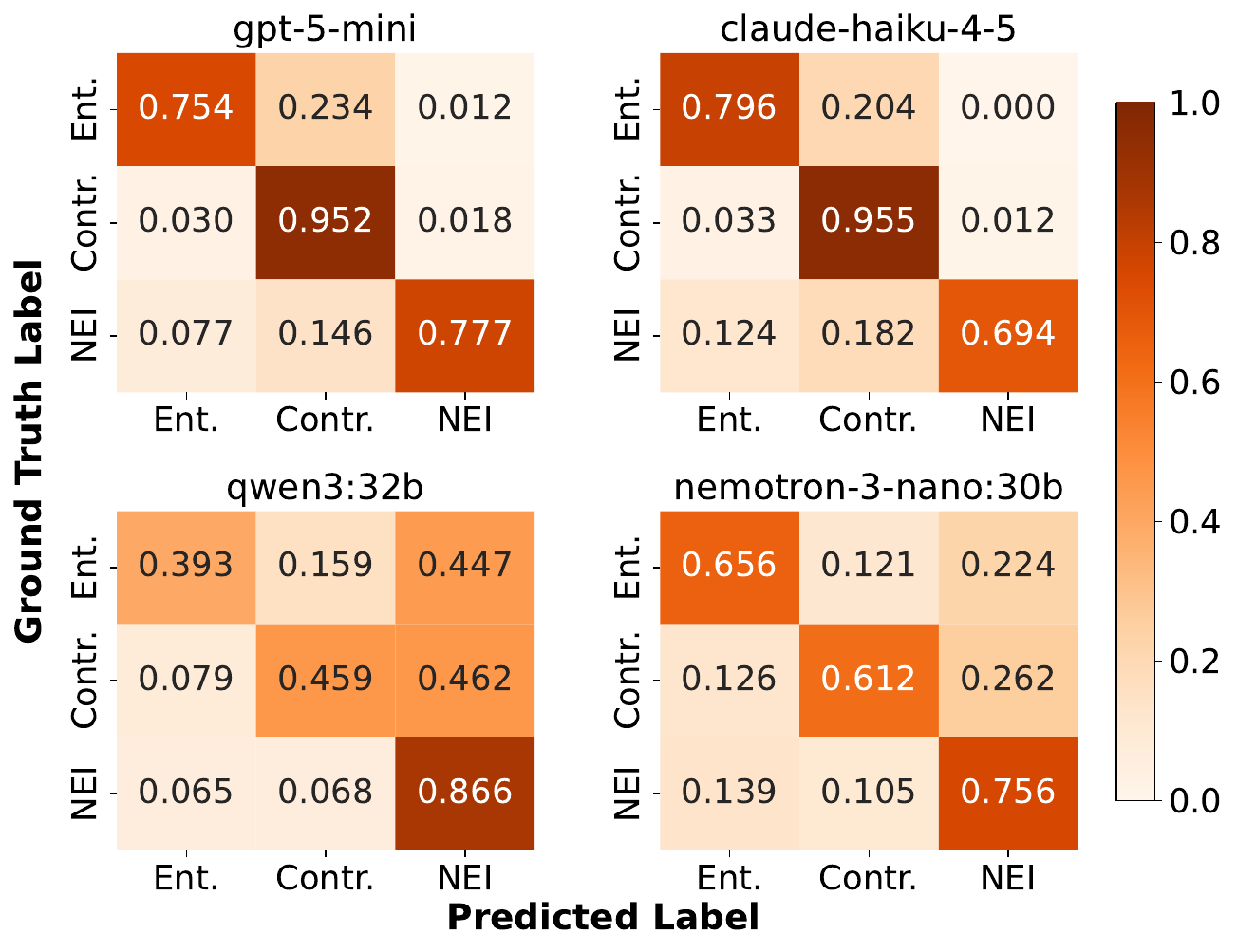}
  \caption{Confusion matrices (normalized) for representative proprietary and open-source models. The primary difference lies in how models handle \textcolor{nei}{NEI}: proprietary models are \emph{biased} against abstention, while open-source models predict it excessively (full results in Figure~\ref{fig:confusion_matrices}).}
  \label{fig:confusion_matrices_selected}
\end{figure}

\vspace{0.5mm}
\noindent \textbf{NEI is Poorly Handled.} We observe that both closed- and open-source models exhibit \emph{opposite but degenerate} behaviors with respect to the \textcolor{nei}{NEI} label. Figure~\ref{fig:confusion_matrices_selected} shows normalized confusion matrices for two top-performing proprietary models and two strong open-source alternatives.

Both \texttt{gpt-5-mini} and \texttt{claude-haiku-4.5} are strongly biased \emph{against} predicting \textcolor{nei}{NEI}. When the ground-truth label is \textcolor{e}{entailed} or \textcolor{c}{contradicted}, they almost never predict \textcolor{nei}{NEI} (close to 0\%), indicating a strong, inherent reluctance. Even when the correct label \emph{is} \textcolor{nei}{NEI}, they hedge heavily, spreading probability mass across the other two labels.

Both \texttt{qwen3} and \texttt{nemotron-3-nano} display the opposite behavior: they over-predict \textcolor{nei}{NEI}. When the ground truth is \textcolor{e}{entailed} or \textcolor{c}{contradicted}, \textcolor{nei}{NEI} is predicted roughly half the time for \texttt{qwen3}, with a milder but similar tendency for \texttt{nemotron-3-nano}.

Notably, the performance gap between closed- and open-source models is driven mostly by how \textcolor{nei}{NEI} is handled. However, neither behavior is desirable: proprietary models effectively avoid abstention, while open-source models over-abstein.

\vspace{1mm}
\noindent \textbf{Non-Superficial Failures.} Our neuro-symbolic baseline operates as a multi-step process driven by SQL queries over the underlying database. Hence, inspecting these queries provides insight into where model failures occur. In Figure~\ref{fig:tool_call_errors}, we plot the distribution of whether executed SQL queries succeed or fail. The two best-performing models, \texttt{gpt-5-mini} and \texttt{claude-haiku-4.5}, achieve SQL query success rates of 93\% and 99\%, respectively. For both models, the remaining failed queries are evenly distributed across correct and incorrect predictions. This suggests that, for stronger models, failures are not superficial (e.g., syntax), but instead reflect breakdowns in reasoning—specifically, in how the model composes queries to navigate the database. 



\section{Conclusion}\label{sec:conclusion}

In this work, we introduce \system, a fact-verification benchmark over large-scale structured data. It contains 80 unique databases, each with an average of 11 tables and 4.5M records (roughly 110M tokens when converted to text). Each claim is \emph{grounded} in evidence deliberately composed of millions of records across multiple tables. \system is a clear step toward real-world claim verification---mirroring the various \emph{challenges} of high-stakes fact-checking in our contemporary world.

Lastly, we conduct extensive experiments with 30 state-of-the-art open- and closed-source models, and find that more than half remain below 55\% accuracy. We further uncover systematic failures around \textit{abstention}---the ability to admit there is not enough evidence to decide---raising concerns about the reliability of LLMs in high-stakes verification.


\section{Limitations}

\noindent \textbf{Reliance on BIRD.} Our benchmark is derived from NL-to-SQL pairs in the BIRD benchmark (Section~\ref{sec:bird}). As a result, any annotation error in BIRD directly propagates to \system. For example, if the NL question is \emph{``How old was the oldest mayor of Seattle when elected?''} but the associated SQL query instead computes the answer for \emph{Portland}, any claims generated from this pair would be incorrectly labeled. To mitigate this risk for public evaluation and model development, we construct our \texttt{test} splits exclusively from claims derived from the \texttt{dev} split of BIRD, which has undergone substantial subsequent cleanup by prior work (e.g.,~\citealp{DBLP:conf/acl/WretbladRBAH24, DBLP:conf/kdd/LiuSL0L25}). Importantly, our dependency on BIRD is not irreversible: as additional errors in BIRD are identified, the corresponding derived claims can be removed from \system on-the-fly.

\vspace{1mm}
\noindent \textbf{Single Evidence Modality.} \system deliberately focuses on structured data as its primary evidence source (which may include numerical, categorical, temporal, and short textual fields). While real-world fact verification often involves multi-modal evidence---such as free-form text, reports, news articles, charts, and other unstructured sources---structured data remains a critical yet comparatively underexplored modality in existing benchmarks. By isolating this setting, \system enables targeted evaluation of reasoning over complex structured evidence. We acknowledge that this focus alone does not capture the full breadth of real-world verification, and view extension to multi-modal evidence as a natural future direction.


\vspace{1mm}
\noindent \textbf{Snapshot Validity and Evaluation Scope.} Entailment, contradiction, and NEI are assessed with respect to the fixed database snapshot used to construct the benchmark. Because the underlying databases come from established public dataset releases---and are therefore static and potentially dated---answers may differ if a system incorporates newer external information. For example, a query about the \textit{``bottom-5 enrollment schools''} could change, even if the claim was correct in the snapshot used for claim generation. As with many ML benchmarks, evaluation should be interpreted relative to the provided database state~\cite{DBLP:conf/nips/LiHQYLLWQGHZ0LC23} rather than a live, continuously updated world.

\vspace{1mm}
\noindent \textbf{Evaluation Solely on SQL.} Our evaluation uses agents equipped with a single tool: the ability to execute SQL queries. Performance therefore depends not only on a model's reasoning ability, but also on its familiarity with SQL. One of the reasons we choose SQL is because, as a language, it makes the required data operations---such as filtering, grouping, aggregation, and ordering---explicit, isolating the core reasoning challenge. But, some models may perform worse than others simply because they are worse at writing SQL (e.g., due to their pre-training data).

\section{Ethical Considerations}

\noindent \textbf{Data Licensing.} \system is constructed from NL-to-SQL pairs derived from the BIRD benchmark, which is released under the CC BY-SA 4.0 license. Accordingly, the \system dataset is also released under the same license, in compliance with the original license terms.



\vspace{1mm}
\noindent \textbf{Privacy}. We do not anonymize any part of \system. \system includes \textit{contradicted} claims, which are synthetically generated by perturbing facts and are explicitly labeled as false within the benchmark. These claims are not intended to introduce new real-world information about entities, but solely to support controlled evaluation of real-life fact-verification systems.

\section{Acknowledgments}

We thank the anonymous reviewers for their constructive feedback on this paper. This work was supported by NSF III 2507117, NSF SHF 2312195, and NSF IIS 2314527. We  thank Dean Light for many insightful discussions throughout the review process. We also thank the UW Database Group members for helpful feedback and David Alexander for conversations on benchmark performance.

\bibliography{references}

\FloatBarrier

\appendix

\section{Tokens Per Database}\label{appendix:db-to-tokens}

In this section, we estimate the number of tokens required to represent a database in context. This analysis supports one of our central claims: in \system, the context of each claim (i.e., its underlying database) exceeds the context window of modern LLMs, which forces solutions to reason in executable programs (e.g., SQL).

To obtain this estimate, we take a straightforward approach. Each database in \system consists of multiple tables. For each database, we export all of its tables as CSV files (header row followed by all data rows), concatenate them into a single text file, and count the resulting tokens using the tokenizer of the \texttt{gpt-5} series. We use this tokenizer as a representative modern LLM tokenizer. 

\begin{figure*}[htbp]
  \includegraphics[width=\textwidth]{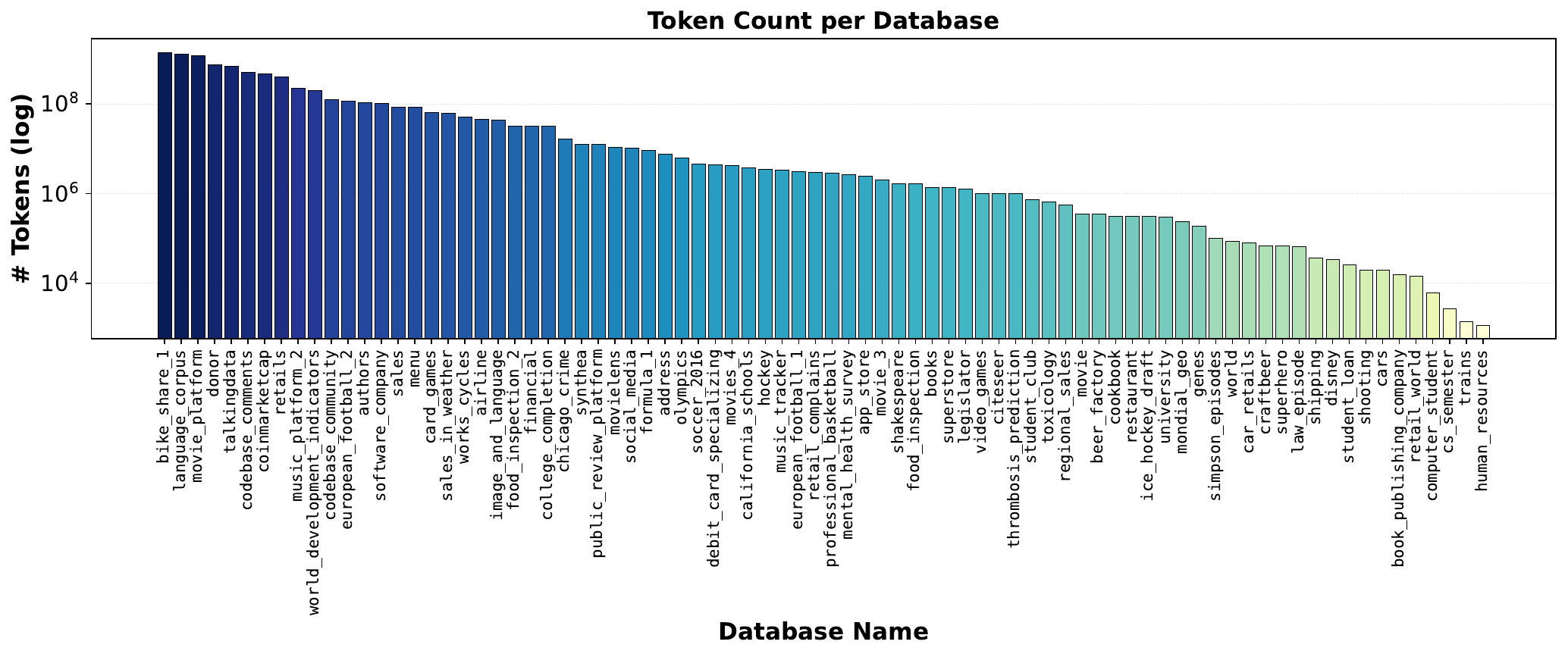}
  \caption{Token count per database (log scale), computed using the tokenizer of the \texttt{gpt-5} series. Each bar corresponds to one of the 80 databases in \system. For more than half of the  databases, naive in-context ``reading'' of the full evidence is practically impossible. Notably, the reported average of 110M tokens of context per claim in Table~\ref{tab:benchmark_stats} reflects the fact that claims are unevenly distributed across databases. In particular, we end up with a lot more claims based on larger databases which means the per-claim token average is higher than a simple average over the 80 databases.}
  \label{fig:tokens_per_db_histogram}
\end{figure*}

Importantly, this estimate is conservative for two reasons. First, it does not encode the relationships and constraints present in the database (e.g., keys, foreign keys, and other schema-level structure). Second, providing structured data in context to an LLM typically requires an explicit representation---such as JSON (\citealt{singha2023tabular})---that clearly associates each value with its column and preserves structure. This additional encoding would substantially increase the token count. The reported numbers should therefore be interpreted as a lower bound. Figure~\ref{fig:tokens_per_db_histogram} shows the exact token counts for each of the 80 databases.

\section{Evaluation}

\subsection{Evaluation Setup}
Table~\ref{tab:hyperparams} illustrates the hyperparameters of the evaluation setup.

\begin{table}[t]
\centering
\small
\setlength{\tabcolsep}{3.5pt}
\begin{tabular}{l c c c c c}
\toprule
\textbf{Model} &
\makecell[c]{\textbf{Context}\\\textbf{Window}} &
\makecell[c]{\textbf{temp.}} &
\makecell[c]{\textbf{topK}} &
\makecell[c]{\textbf{topP}} \\
\midrule
\texttt{gpt-4o-mini}     & 128K      & 1.0  & -- & 1.0 \\
\texttt{gpt-4.1-nano}     & 1M     & 1.0  & -- & 1.0 \\
\texttt{gpt-5-nano}     & 400K      & 1.0  & -- & 1.0 \\
\texttt{gpt-5-mini}     & 400K      & 1.0  & -- & 1.0 \\
\texttt{gpt-oss:20b}     & 128K      & 1.0  & -- & 1.0 \\
\midrule
\texttt{gemini-2.5-flash}     & 1M    & 1.0  & 64 & 0.95 \\
\texttt{gemini-3-flash}     & 1M    & 1.0  & 64 & 0.95 \\
\midrule
\texttt{claude-3-haiku}     & 200K    & 1.0  & -- & -- \\
\texttt{claude-3-5-haiku}     & 200K    & 1.0  & -- & -- \\
\texttt{claude-haiku-4-5}     & 200K    & 1.0  & -- & -- \\
\midrule
\texttt{qwen3:1.7b}     & 40K    & 0.6  & 20 & 0.95 \\
\texttt{qwen3:4b}     & 246K      & 0.6  & 20 & 0.95 \\
\texttt{qwen3:8b}    & 40K        & 0.6  & 20 & 0.95 \\
\texttt{qwen3:14b}    & 40K       & 0.6  & 20 & 0.95 \\
\texttt{qwen3:32b}     & 40K      & 0.6  & 20 & 0.95 \\
\texttt{qwen3-coder:30b}  & 256K   & 0.7  & 20 & 0.8  \\
\texttt{qwq:32b}  & 40K & 1  & 40 & 0.95  \\
\midrule
\texttt{mistral-nemo:12b}   & 1M   & 0.0   & -- & --   \\
\texttt{ministral-3:3b}   & 256K   & 0.15 & 50 & 90   \\
\texttt{ministral-3:8b}  & 256K     & 0.15 & 50 & 90   \\
\texttt{ministral-3:14b}  & 256K     & 0.15 & 50 & 90   \\
\texttt{mistral-small:22b} & 128K     & 0.0  & -- & --   \\
\texttt{magistral:24b}   & 39K     & 0.7  & -- & 0.95 \\
\texttt{devstral:24b}   & 128K     & 0.0   & -- & --   \\
\texttt{devstral-small-2:24b} & 384K   & 0.15 & -- & --   \\
\midrule
\texttt{nemotron-3-nano}  & 1M    & 1.0  & -- & 1.0  \\
\midrule
\texttt{llama3.2:3b}     & 128K & 0.0  & -- & -- \\
\texttt{llama3.1:8b}     & 128K & 0.0  & -- & -- \\
\midrule
\texttt{cogito:14b}      & 128K & 0.0  & -- & -- \\
\texttt{cogito:32b}      & 128K & 0.0  & -- & -- \\
\bottomrule
\end{tabular}
\caption{Extra information and hyperparameters. Unavailable or N/A values are shown as ``--''. Numbers also come from HuggingFace and Ollama.}
\label{tab:hyperparams}
\end{table}

\subsection{Structured Outputs}\label{appendix:structured}

A few years ago, getting \textit{structured outputs}---responses that adhere to strict user-defined JSON schemas---back from LLMs was a painful process. Making it work would typically require careful prompting and repeated retries to get outputs in the right format. At the same time, this feature was becoming increasingly important for any serious application that relied on LLMs.

Thankfully, the situation has improved substantially. Most major AI providers (e.g., OpenAI, Anthropic, Google) now support some form of \emph{structured outputs}. There are multiple ways to implement this, but one of the most reliable is grammar-constrained decoding~\cite{DBLP:conf/emnlp/GengJP023}. In this approach, the JSON schema is compiled into a context-free grammar, and during decoding the model's logits are masked so that only grammar-valid tokens can be generated.

Concretely, this means that when a JSON schema is provided, the model is \emph{guaranteed} to produce a syntactically valid response. For example, if we request an \texttt{enum} over three possible verdicts (\textcolor{e}{entailed}, \textcolor{c}{contradicted}, \textcolor{nei}{NEI}), the output is guaranteed to be one of these values. This greatly simplifies evaluation and removes an entire class of failure modes unrelated to model reasoning.

This level of control is only possible because model providers have direct access to their own decoding process. As a result, most proprietary APIs expose structured output functionality directly. However, when using agentic frameworks (e.g., OpenAI Agent SDK, Pydantic AI, Google ADK, etc.) that connect to third-party or open-source models through adapters such as LiteLLM, this guarantee often breaks down.

For example, some leading agentic frameworks simulate structured outputs by injecting a prompt that forces the model to call a predefined tool for its final answer\footnote{\scriptsize
\url{https://github.com/pydantic/pydantic-ai/issues/242},
\url{https://github.com/openai/openai-agents-python/issues/1778\#issuecomment-3316092585}
}, rather than enforcing the schema at decoding time when the Grammar Constraint Decoding (GCD) option is available (e.g., on open-source models). Aside from bloating the model context, this approach is inherently brittle: the LLM can simply ignore the instruction.

This creates an evaluation asymmetry. Models without guaranteed structured decoding may fail due to formatting errors even when their underlying prediction is correct, while models with GCD will always produce a valid label (which implies a $33\%$ accuracy floor). Since our goal is not to benchmark how well models, vendors and agentic frameworks support structured outputs, but rather how well can models reason over large data, directly penalizing such failures would be unfair.

Our policy is therefore simple. For models where structured outputs are not guaranteed---typically due to limitations of the agentic framework we use---we re-run the test whenever a response violates the expected schema. This occurs in only a small fraction of cases and is usually resolved with one or two re-runs.

\begin{figure}[t]
    \centering
  \includegraphics[width=0.99\columnwidth]{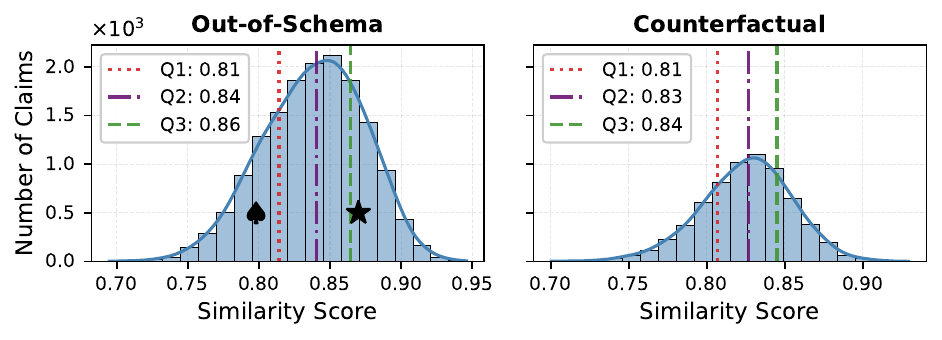}
  \caption{Distribution of similarity scores between generated claims and their gold context for both \textit{counterfactual} and \textit{out-of-schema} claims. These claims can ``drift'' from the database concepts so we embed them and measure semantic similarity between them and the also embedded golden context. Higher scores indicate that claims stay ``closer'' to the underlying data concepts. The two highlighted claim examples ($\clubsuit$, $\bigstar$) are discussed in Section~\ref{sec:embeddings}.}
  \label{fig:similarity_appendix}
\end{figure}

 \begin{figure*}[t]
    \centering
  \includegraphics[width=0.99\textwidth]{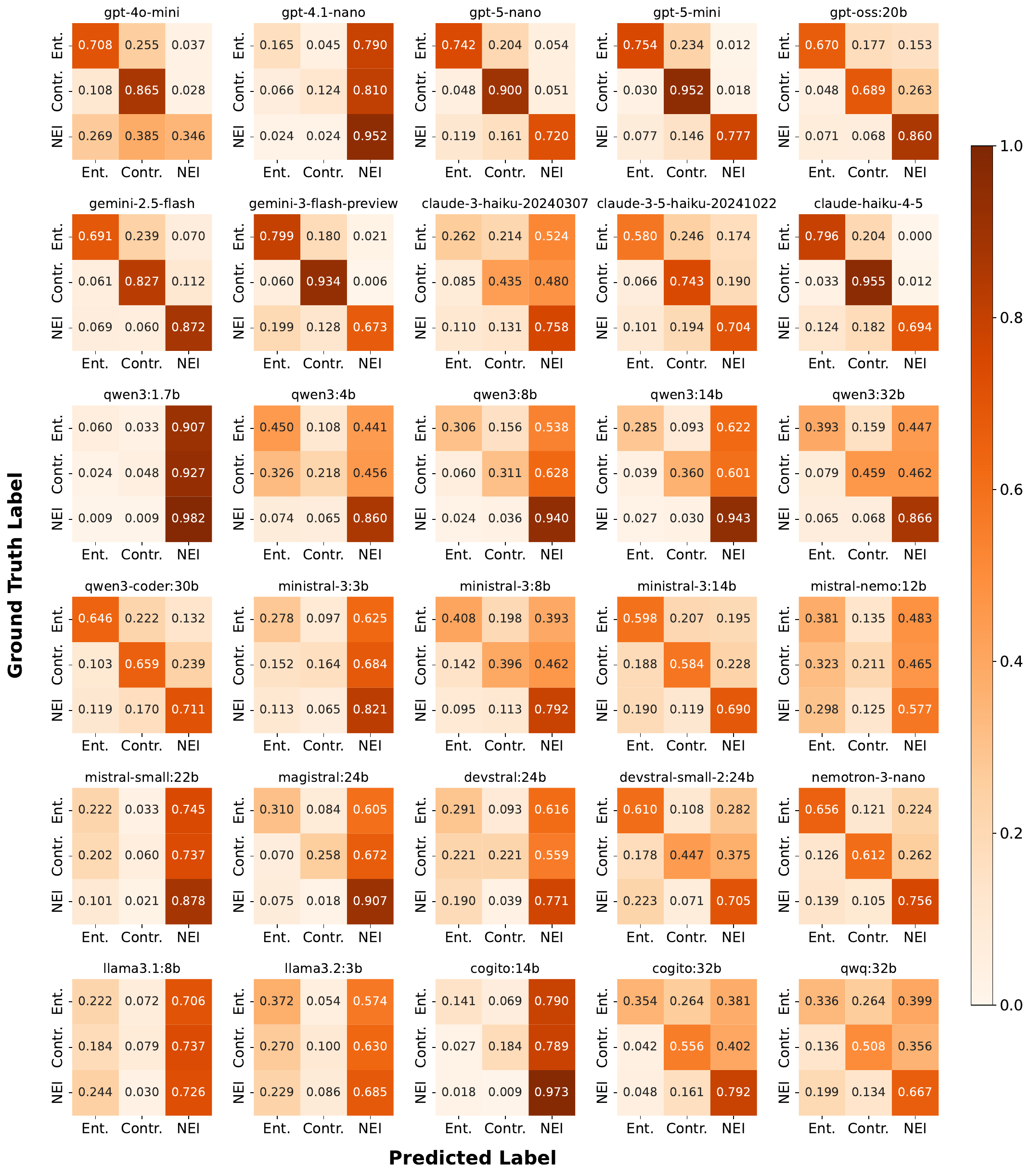}
  \caption{Public \texttt{test} set. Confusion matrices (normalized) for all models. One takeaway is that the primary difference in top-performing open- and closed-source models lies in how they handle \textcolor{nei}{NEI}: proprietary models are \emph{biased} against abstention, while open-source models predict it excessively.}
  \label{fig:confusion_matrices}
\end{figure*}

\begin{figure*}[t]
  \includegraphics[width=\textwidth]{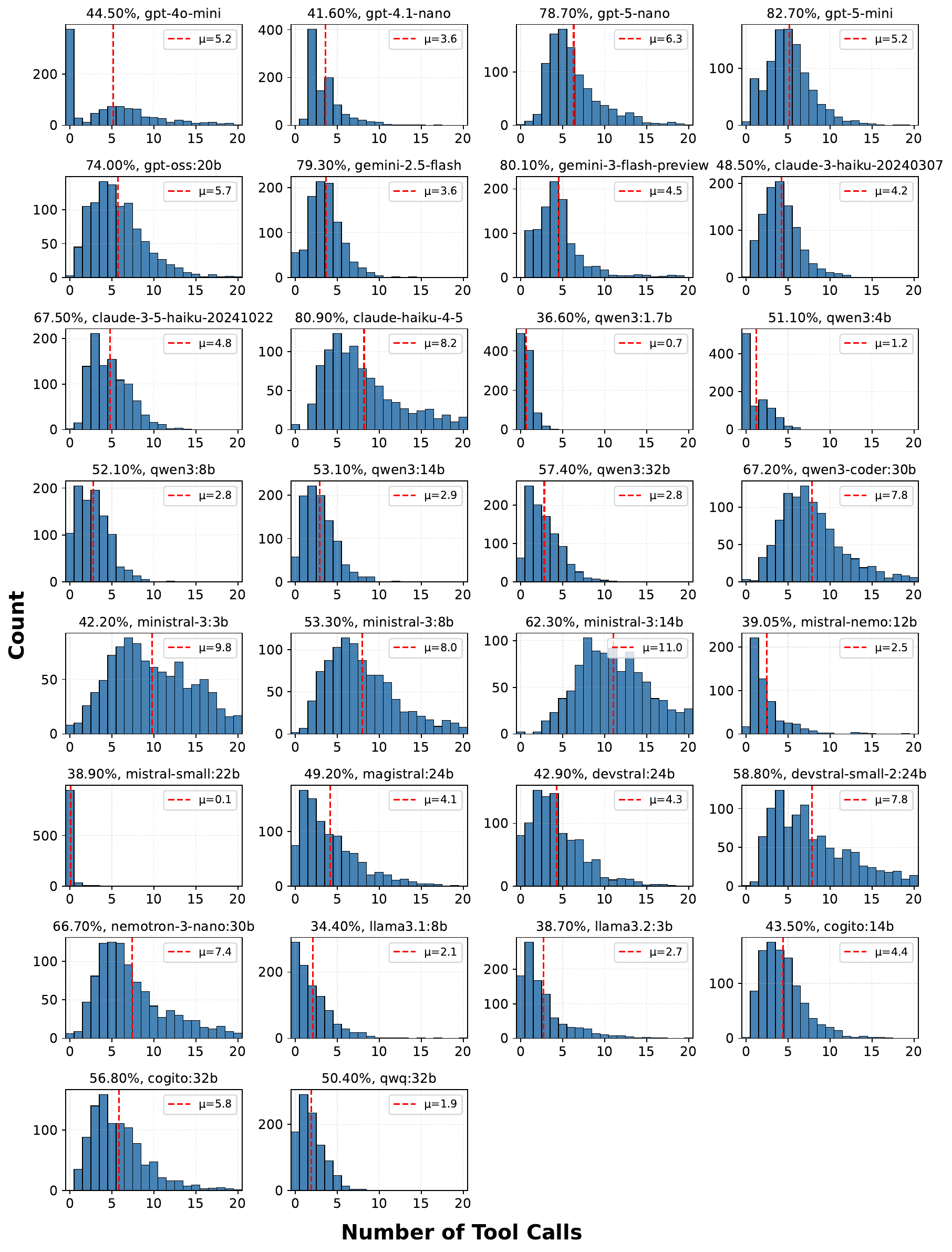}
  \caption{Public \texttt{test} set. Distribution of the number of tool calls for each model on the 1,000 examples of the public \texttt{test} set of \system. The best-performing model is \texttt{gpt-5-mini} (0.83 acc.) which gives us insight on \textit{how the distribution of tools calls} should probably look like for a model to do well on our benchmark. Some models often decide not to use a tool call at all, which is obviously a losing strategy as they have no idea of the underlying data environment to answer correctly. We try to avoid this by ``forcing'' the model to inspect the data first, by including instructions like \textit{``Use the available tools to query the database and gather evidence before making a decision''} and \textit{``You should always start by querying the database for the schema''} (Figure~\ref{fig:verifier_prompt}) which are ignored.}
  \label{fig:tool_calls_histogram}
\end{figure*}

\section{Prompts}

In Figures~\ref{fig:contradicted_prompt}, \ref{fig:entailed_prompt} and \ref{fig:nei_prompt} we showcase the prompts used in creating the claims (Section~\ref{sec:claim_gen}). In Figures~\ref{fig:con_entailed_judge_prompt} and \ref{fig:nei_judge_prompt} we illustrate the two prompts used for the judging process---one for \textcolor{c}{contradicted} and \textcolor{e}{entailed} claims and one for \textcolor{nei}{NEI} since the latter has extra rubrics and extra golden context (Section~\ref{sec:judges}). Lastly, in Figure~\ref{fig:verifier_prompt}, we show the single, optimized verifier prompt for all models in our evaluations.

\section{SQL Queries and ASTs}\label{sec:appendix_qast}

\begin{figure*}[htbp]
    \centering
    \begin{minipage}{0.5\textwidth}
    \begin{minted}[gobble=4, frame=single]{sql}
    SELECT business_id 
    FROM Business 
    WHERE state LIKE 'AZ' AND stars = 5
    \end{minted}
    \end{minipage}
    \includegraphics[width=0.5\textwidth]{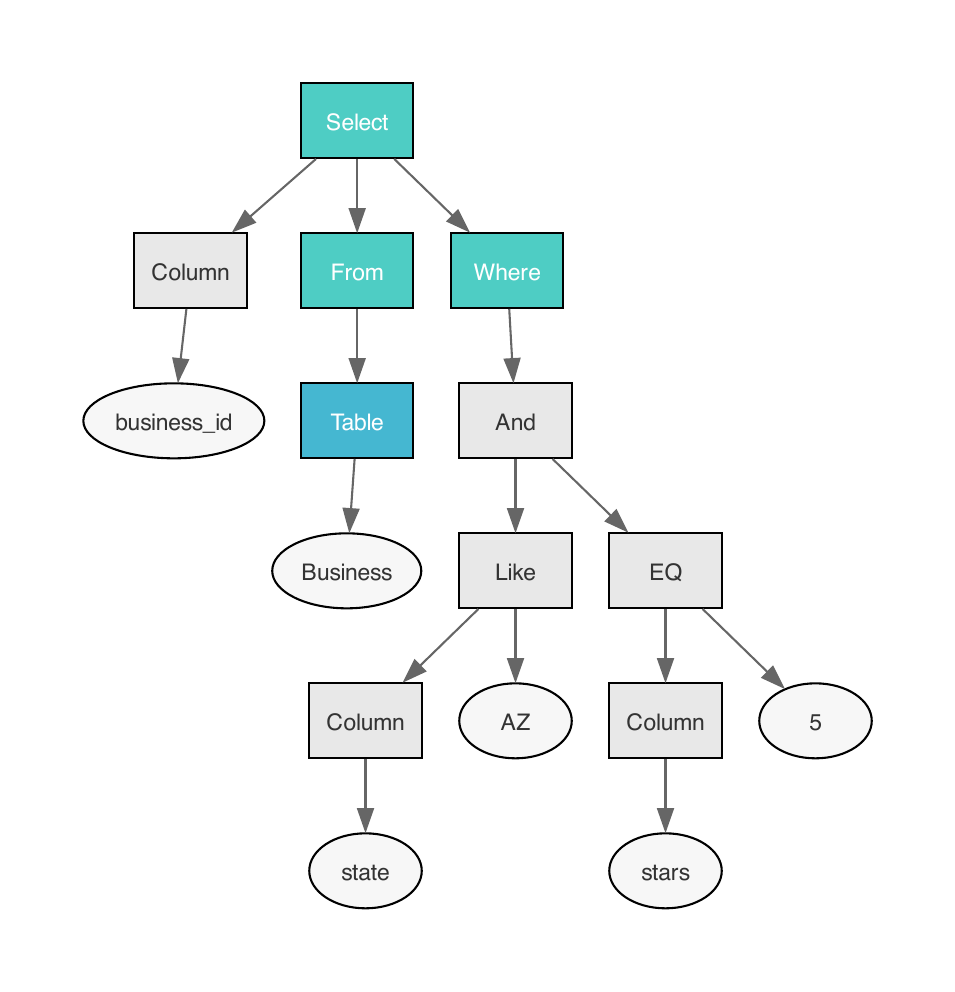}
  \caption{\textbf{\ding{55} Not Compositional}. The SQL query (top) and its AST (bottom). The query is excluded since it only applies local filters (\texttt{state}, \texttt{stars}) and does not involve aggregations, joins, or other operations that combine information across many records.}
  \label{fig:q7}
\end{figure*}

\begin{figure*}[htbp]
    \centering
    \begin{minipage}{0.6\textwidth}
    \begin{minted}[gobble=4, frame=single]{sql}
    SELECT COUNT(T1.inspection_id) 
    FROM inspection AS T1 INNER JOIN employee AS T2 
        ON T1.employee_id = T2.employee_id 
    WHERE T2.first_name = 'Lisa' AND    
        T2.last_name = 'Tillman' AND 
        T1.results = 'Out of Business'
    \end{minted}
    \end{minipage}
    \includegraphics[width=\textwidth]{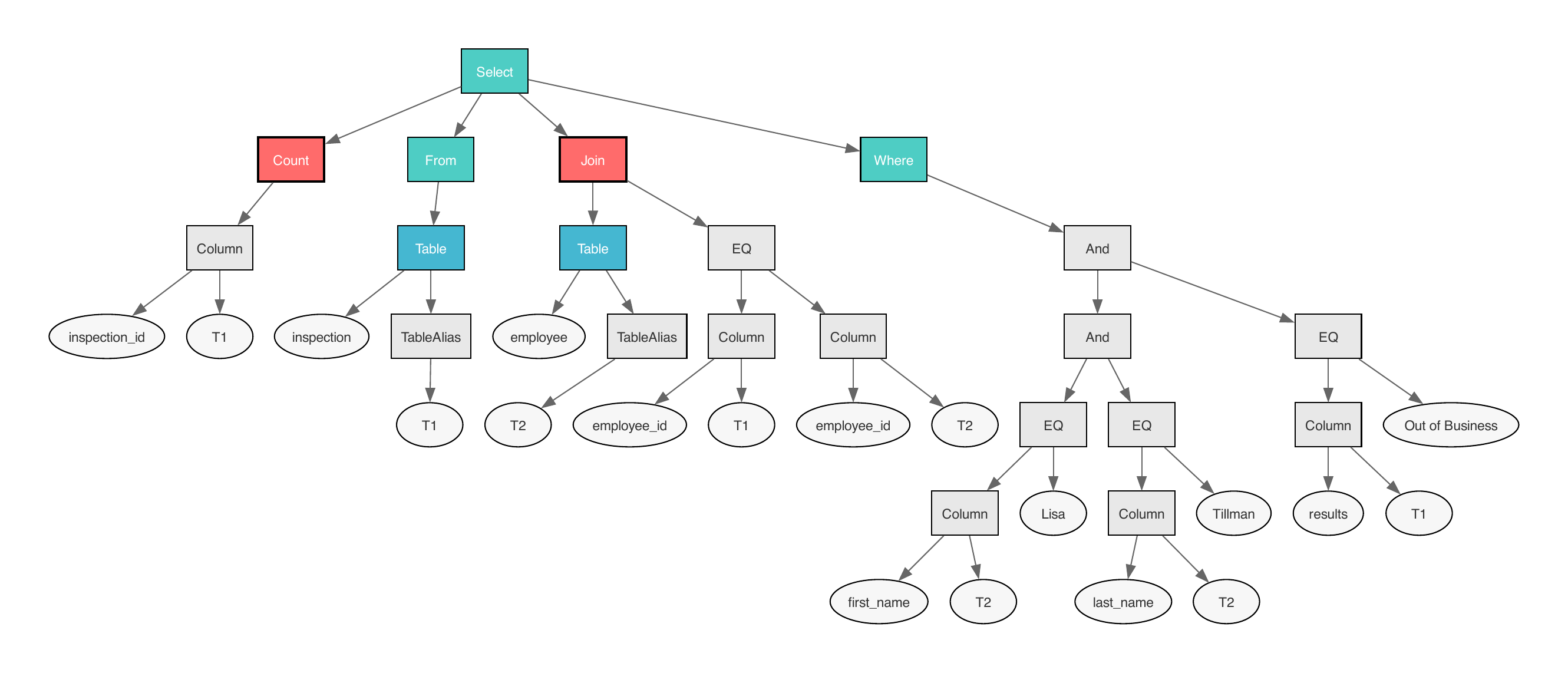}
  \caption{\textbf{\ding{51} Compositional}. The SQL query (top) and its AST (bottom). The query is retained since its AST contains an aggregate function (\texttt{COUNT}), which counts a large number of records and collapses them into a single value (the count). Verifying a claim derived from this query therefore requires compositional reasoning over many records, which cannot be performed by naively ``reading'' the records in-context and counting by an LLM alone.}
  \label{fig:q4}
\end{figure*}

\begin{figure*}[htbp]
    \centering
    \begin{minipage}{0.55\textwidth}
    \begin{minted}[gobble=4, frame=single]{sql}
    SELECT T2.movie_title, T1.user_id,
        T1.rating_score, T1.critic 
    FROM ratings AS T1 INNER JOIN movies AS T2 
        ON T1.movie_id = T2.movie_id 
    WHERE T1.critic IS NOT NULL
    \end{minted}
    \end{minipage}
    \includegraphics[width=\textwidth]{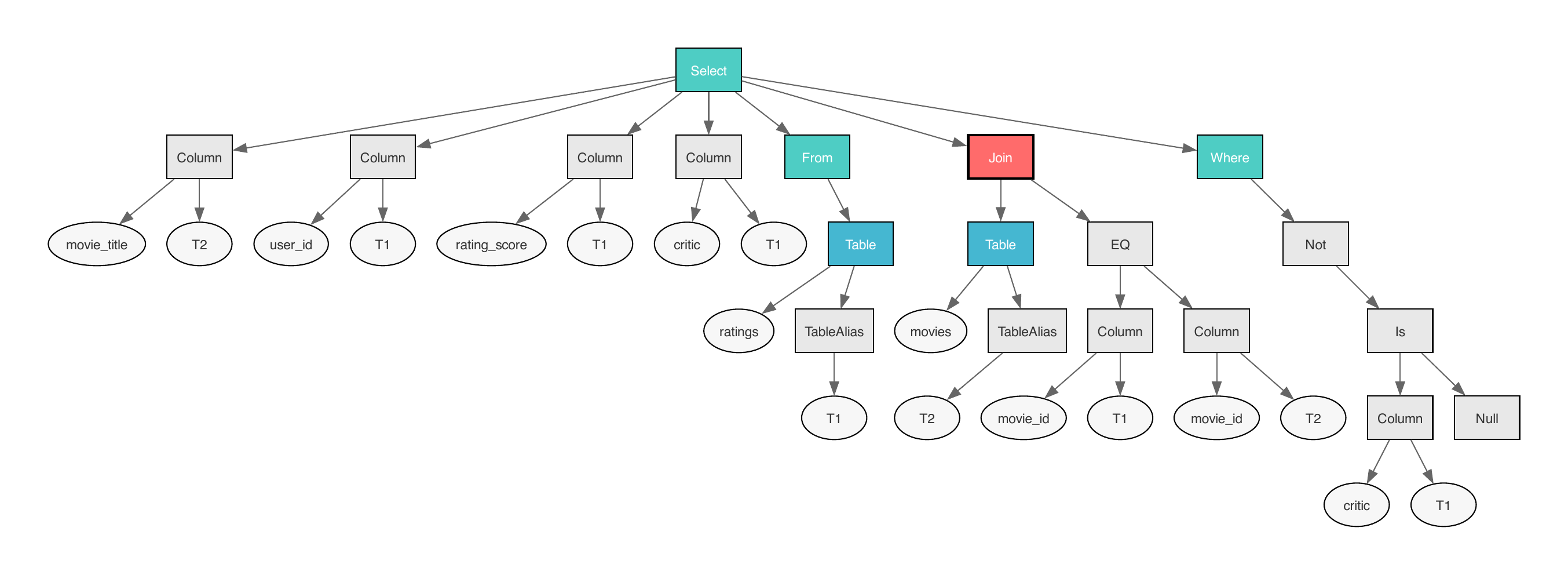}
  \caption{\textbf{\ding{55} Not Compositional}. The SQL query (top) and its AST (bottom). The query is excluded since it does not contain aggregations, orderings, window functions, or joins involving \emph{three or more} tables (see Section~\ref{sec:filtering}).}
  \label{fig:q8}
\end{figure*}

\begin{figure*}[htbp]
    \centering
    \begin{minipage}{0.55\textwidth}
    \begin{minted}[gobble=4, frame=single]{sql}
    SELECT T1.title
    FROM movie AS T1
        INNER JOIN movie_keywords AS T2
            ON T1.movie_id = T2.movie_id
        INNER JOIN keyword AS T3
            ON T2.keyword_id = T3.keyword_id
    WHERE T3.keyword_name = 'extremis' 
    \end{minted}
    \end{minipage}
    \includegraphics[width=\textwidth]{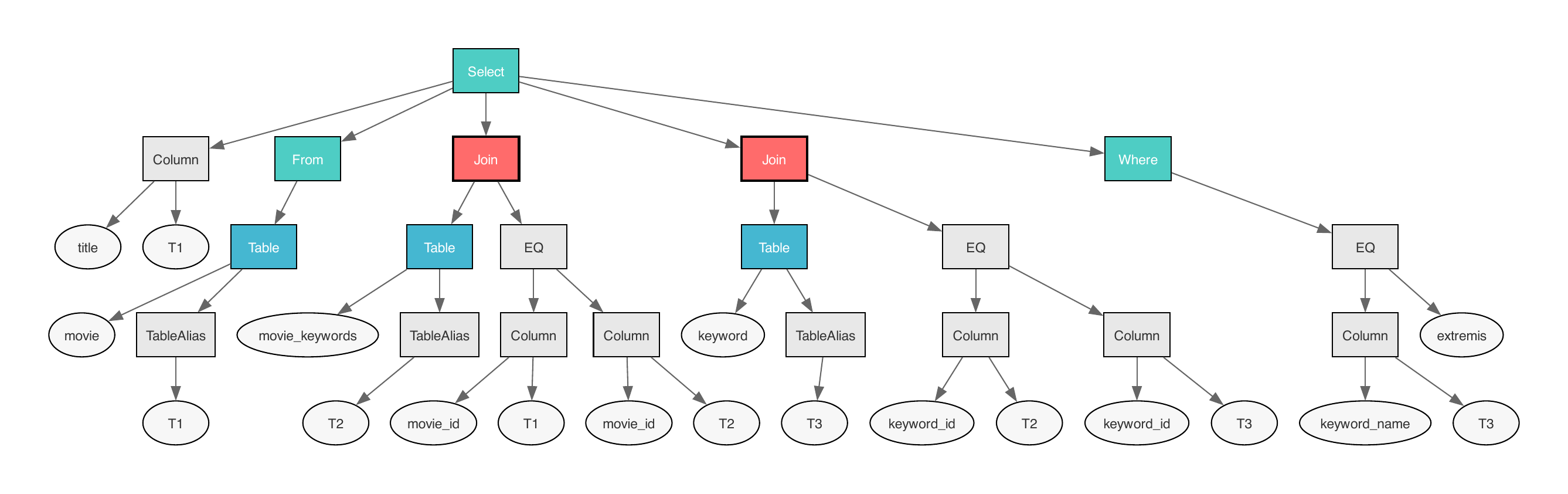}
  \caption{\textbf{\ding{51} Compositional}. The SQL query (top) and its AST (bottom). The query is retained since it joins three tables together---the tables \texttt{movie} aliased as \texttt{T1}, \texttt{movie\_keywords} aliased as \texttt{T2}, and \texttt{keyword} aliased as \texttt{T3}---which is a compositional operation as defined in Section~\ref{sec:filtering}. A verifier must explicitly combine these three tables in order to verify a claim derived from this SQL query.}
  \label{fig:q9}
\end{figure*}

This section provides concrete examples of SQL queries together with their corresponding abstract syntax trees (ASTs). Figures~\ref{fig:q7}--\ref{fig:q9} illustrate how the filtering rules described in Section~\ref{sec:filtering} are applied in practice.

 \section{Similarity Distributions}
 We provide the \textit{counterfactual} and \textit{out-of-schema} similarity distributions (Section~\ref{sec:embeddings}) in Figure~\ref{fig:similarity_appendix}.

\begin{figure}[t]
\centering
\begin{minted}[fontsize=\tiny, breaklines, frame=single]{python}
from pydantic import BaseModel, Field

class ContradictedClaim(BaseModel):
    contradicted_claim: str = Field(
        ...,
        description=(
            "A contradicted claim."
        )
    )

class ContradictedClaimCollection(BaseModel):
    collection: list[ContradictedClaim]
\end{minted}
\begin{minted}[fontsize=\tiny, gobble=4, frame=single, breaklines=true, breaksymbol={}]{markdown}
    ## Role
    You are a **misleading** spokesperson  **in a controlled evaluation setting**.
    
    ## Task
    Given the following inputs:
    - A question
    - Its correct answer
    - The data domain
    - Optional external knowledge (clarifications)
    
    Your task is to produce natural language claims that are factually incompatible with the provided answer. In other words, any reader who knows the correct answer would judge your claim to be false.
    
    ## Requirements
    - Each claim must be self-contained and must not use opaque references to earlier context (e.g., "the answer," "the question," "the earlier claim", etc.). Instead, any needed context should be stated explicitly within each claim.
    - Each claim must contradict or be factually incompatible with the answer, directly or indirectly.
    - Do not restate or explain the external knowledge; assume it is already known to the reader.
    - Produce between 1 and 3 claims.
    
    ## Example
    
    ### Input
    {
      "question": "Which three districts recorded the highest graduation rates in 2022?",
      "answer": [
        {
          "DistrictName": "Redwood Coast Unified",
          "GradRate": 0.97
        },
        {
          "DistrictName": "Sierra Vista Union",
          "GradRate": 0.96
        },
        {
          "DistrictName": "Mission Creek Unified",
          "GradRate": 0.95
        }
      ],
      "domain": "California Schools",
      "external-knowledge": "GradRate = Number of graduates / Total number of eligible seniors"
    }
    
    ### Output
    Redwood Coast Unified did not lead California's graduation rankings in 2022 — it was Riverbend Joint Unified that posted the top rate.
    
    Sierra Vista Union is no longer among the highest graduation-rate districts in 2022.
    
    Fairmont Hills Unified surpassed Redwood Coast Unified with 98% of its eligible seniors graduating in 2022, according to data in CA.
\end{minted}
\caption{Prompt (1-shot) for generating \textcolor{c}{contradicted} claims. It is highly optimized for \texttt{gpt-5} specifically, by following the work of \citet{Kotha-2025-OpenAI-Prompting-Guide}. The \textit{output schema} is provided here as python code as we do not know how OpenAI encodes it in the prompt internally. Effectively, the model returns a JSON object that adheres to \textcolor{c}{\texttt{ContradictedClaimCollection}}, i.e., a collection of \textcolor{c}{contradicted} claims. The number of generated claims usually stays between 1-3 as per the instructions.}
\label{fig:contradicted_prompt}
\end{figure}

\begin{figure}[t]
\centering
\begin{minted}[fontsize=\tiny, breaklines, frame=single]{python}
from pydantic import BaseModel, Field

class EntailedClaim(BaseModel):
    entailed_claim: str = Field(
        ...,
        description=(
            "An entailed claim."
        )
    )

class EntailedClaimCollection(BaseModel):
    collection: list[EntailedClaim]
\end{minted}
\begin{minted}[fontsize=\tiny, gobble=4, frame=single, breaklines=true, breaksymbol={}]{markdown}
    ## Role
    You are an **honest** spokesperson **in a controlled evaluation setting**.
    
    ## Task
    Given the following inputs:
    - A question
    - Its correct answer
    - The data domain
    - Optional external knowledge (clarifications)
    
    Your task is to produce natural language claims that are consistent with and supported by the provided answer. In other words, any reader who knows the correct answer would judge your claim to be true.
    
    ## Requirements
    - Each claim must be self-contained and must not use opaque references to earlier context (e.g., "the answer," "the question," "the earlier claim", etc.). Instead, any needed context should be stated explicitly within each claim.
    - Each claim must follow from or be fully supported by the answer, directly or indirectly.
    - Do not restate or explain the external knowledge; assume it is already known to the reader.
    - Produce between 1 and 3 claims.
    
    ## Example
    
    ### Input
    {
      "question": "Which three districts recorded the highest graduation rates in 2022?",
      "answer": [
        {
          "DistrictName": "Redwood Coast Unified",
          "GradRate": 0.97
        },
        {
          "DistrictName": "Sierra Vista Union",
          "GradRate": 0.96
        },
        {
          "DistrictName": "Mission Creek Unified",
          "GradRate": 0.95
        }
      ],
      "domain": "California Schools",
      "external-knowledge": "GradRate = Number of graduates / Total number of eligible seniors"
    }
    
    ### Output
    Redwood Coast Unified led California's graduation rankings in 2022 with a 97% rate.
    
    In 2022, California's strongest graduation results came from Redwood Coast Unified, which saw 97% of its eligible seniors finish high school. Sierra Vista Union and Mission Creek Unified followed closely, with graduation rates of 96% and 95%, respectively.
    
    Mission Creek Unified achieved a graduation rate of 95% in 2022, placing it among California's top three districts. It ranked just behind Redwood Coast Unified and Sierra Vista Union.
\end{minted}
\caption{Prompt (1-shot) for generating \textcolor{e}{entailed} claims. It is highly optimized for \texttt{gpt-5} specifically, by following the work of \citet{Kotha-2025-OpenAI-Prompting-Guide}. The \textit{output schema} is provided here as python code as we do not know how OpenAI encodes it in the prompt internally. Effectively, the model returns a JSON object that adheres to \textcolor{e}{\texttt{EntailedClaimCollection}}, i.e., a collection of \textcolor{e}{entailed} claims. The number of generated claims usually stays between 1-3 as per the instructions.}
\label{fig:entailed_prompt}
\end{figure}

\begin{figure}[t]
\centering
\begin{minted}[fontsize=\tiny, breaklines, frame=single]{python}
from pydantic import BaseModel, Field
from typing import Literal

class NoInfoClaim(BaseModel):
    no_info_claim: str = Field(
        ...,
        description="A NOT ENOUGH INFO claim."
    )
    category: Literal[
        "Out-of-Schema", "Subjective", "Counterfactual"
    ] = Field(
        ...,
        description="The category of the NOT ENOUGH INFO claim."
    )

class NoInfoClaimCollection(BaseModel):
    collection: list[NoInfoClaim]
\end{minted}
\begin{minted}[fontsize=\tiny, gobble=4, frame=single, breaklines=true, breaksymbol={}]{markdown}
    ## Role
    You are a neutral spokespearson **in a controlled evaluation setting**.
    
    ## Task
    Given the following inputs:
    - A question
    - Its correct answer
    - The data domain
    - The schema of the database
    - Optional external knowledge (clarifications)
    
    Your task is to produce natural language claims whose truth **cannot** be determined from the database or the given Q/A. That is, even with full access to both the database and the correct answer, these claims cannot be definitively verified or falsified.
    
    ## Requirements
    - Each claim must be self-contained and must not use opaque references to earlier context (e.g., "the answer," "the question," "the earlier claim", etc.). Instead, any needed context should be stated explicitly within each claim.
    - Each claim must *not* be entailed or contradicted by the answer, directly or indirectly.
    - Each claim must fall into at least one of these categories:
      1. **Out-of-schema** — involves concepts the database doesn't store or represent anywhere in its schema.
      2. **Subjective/evaluative** — expresses opinions or judgments that cannot be objectively verified.
      3. **Counterfactual/hypothetical** — describes an imagined or "what if" situation that is not reflected in the actual data.
    - Produce between 1 and 5 claims.
    - Do not restate or explain the external knowledge; assume it is already known to the reader.
\end{minted}
\caption{Prompt (zero-shot) for generating \textcolor{nei}{NEI} claims. It is highly optimized for \texttt{gpt-5} specifically, by following the work of \cite{Kotha-2025-OpenAI-Prompting-Guide}. The \textit{output schema} is provided here as python code as we do not know how OpenAI encodes it in the prompt internally. Effectively, the model returns a JSON object that adheres to \textcolor{nei}{\texttt{NoInfoClaimCollection}}, i.e., a collection of  \textcolor{nei}{NEI} claims. The number of generated claims usually stays between 1-5 as per the instructions.}
\label{fig:nei_prompt}
\end{figure}

\begin{figure}[t]
\centering
\begin{minted}[fontsize=\tiny, breaklines, frame=single]{python}
from pydantic import BaseModel, Field
from typing import Literal

class ClaimQuality(BaseModel):
    label_correct: Literal["yes", "no"] = Field(
        ..., 
        description=(
            'Is the assigned label of the claim (ENTAILED/CONTRADICTED) '
            'correct given the gold information? Answer "yes" if the '
            'label of the claim follows from the gold information; "no" '
            'otherwise. If you are unsure, answer "no".'
        )
    )
    free_of_meta_references: Literal["yes", "no"] = Field(
        ...,
        description=(
            'Does the claim avoid meta-references to the question, '
            'answer, or prior text (e.g., "this question", '
            '"the answer above", "as mentioned earlier")? Answer '
            '"yes" if it is completely free of meta-references; '
            '"no" otherwise. References to provided external '
            'knowledge do not count as meta-references.'
        )
    )
    reasoning: str = Field(
        ...,
        description=(
            'Brief explanation (1-2 sentences) justifying your '
            'evaluation (especially for "label_correct").'
        )
    )
\end{minted}
\begin{minted}[fontsize=\tiny, gobble=4, frame=single, breaklines=true, breaksymbol={}]{markdown}
    Your task is to evaluate a natural-language claim across two criteria. You will be given a gold context composed of a question, its answer, the domain, and optional external knowledge. Treat the gold context as the authoritative ground truth.
    
    You will be given a claim labeled as either ENTAILED (supported by the gold context) or CONTRADICTED (refuted by the gold context). Using this gold information, assess whether the claim is correctly labeled, and whether it is free of meta-references. More detailed instructions for the two evaluation criteria are provided along with the JSON schema below.
    
    Your answer should be in JSON format, adhering to the following schema:
    <SCHEMA GENERATED HERE>
\end{minted}
\caption{Judge prompt used for \textcolor{c}{contradicted} and \textcolor{e}{entailed} claims.
The prompt consists of two parts: a Pydantic schema that specifies the required JSON output (label correctness, self-containment, and a short justification), and a natural-language instruction block that explains the judging task and the available gold context. The Python block defines the output schema, which is injected into the prompt at runtime in the placeholder \texttt{<SCHEMA GENERATED HERE>}.}
\label{fig:con_entailed_judge_prompt}
\end{figure}

\begin{figure}[t]
\centering
\begin{minted}[fontsize=\tiny, breaklines, frame=single]{python}
from pydantic import BaseModel, Field
from typing import Literal

class NEIClaimQuality(BaseModel):
    label_correct: Literal["yes", "no"] = Field(
        ..., 
        description=(
            'Is the assigned label of the claim (ENTAILED/CONTRADICTED) '
            'correct given the gold information? Answer "yes" if the '
            'label of the claim follows from the gold information; "no" '
            'otherwise. If you are unsure, answer "no".'
        )
    )
    free_of_meta_references: Literal["yes", "no"] = Field(
        ...,
        description=(
            'Does the claim avoid meta-references to the question, '
            'answer, or prior text (e.g., "this question", '
            '"the answer above", "as mentioned earlier")? Answer '
            '"yes" if it is completely free of meta-references; '
            '"no" otherwise. References to provided external '
            'knowledge do not count as meta-references.'
        )
    )
    category_correct: Literal["yes", "no"] = Field(
        ..., 
        description=(
            'Is the assigned category of the claim '
            '(OUT-OF-SCHEMA/COUNTERFACTUAL/SUBJECTIVE) correct given '
            'the gold information? Answer "yes" if the category of the '
            'claim follows from the gold information; "no" otherwise.'
        )
    )
    schema_leakage: Literal["yes", "no"] = Field(
        ...,
        description=(
            'Does the claim expose database schema details or '
            'technical artifacts? Answer "yes" if it exposes '
            'schema details (e.g., table names, column names, etc.); '
            '"no" if it does not.'
        )
    )
    reasoning: str = Field(
        ...,
        description=(
            'Brief explanation (1-2 sentences) justifying your '
            'evaluation (especially for "label_correct" and '
            '"category_correct").'
        )
    )
\end{minted}
\begin{minted}[fontsize=\tiny, gobble=4, frame=single, breaklines=true, breaksymbol={}]{markdown}
    Your task is to evaluate a natural-language claim across several criteria. You will be given a gold context composed of a question, its answer, the domain, optional external knowledge, and the complete database schema underlying the gold context. Treat the gold context as the authoritative ground truth.
    
    You will be given a claim with an assigned NOT ENOUGH INFO (NEI) label, meaning that its truth cannot be determined from the gold context, even with full access to the database. The claim is also assigned an NEI category: OUT-OF-SCHEMA (depends on information not stored in the database), SUBJECTIVE (expresses opinions or judgments), or COUNTERFACTUAL (describes hypothetical scenarios). Using the gold context, assess whether the NEI label and category are correct, whether the claim is free of meta-references, and whether it leaks schema details of the database. More detailed descriptions for each evaluation criterion are provided in the JSON schema below.
    
    Your answer should be in JSON format, adhering to the following schema:
    <SCHEMA GENERATED HERE>
\end{minted}
\caption{Judge prompt used for \textcolor{nei}{NEI} claims.
Compared to the prompt for entailed and contradicted claims, this prompt includes additional evaluation criteria specific to NEI cases. In addition to label correctness and self-containment, judges assess whether the assigned NEI category is correct and whether the claim leaks database schema details. The instruction block also gives judges access to the full database schema. The Python block defines the output schema, which is injected into the prompt at runtime in the placeholder \texttt{<SCHEMA GENERATED HERE>}.}
\label{fig:nei_judge_prompt}
\end{figure}

\begin{figure}[t]
\centering
\begin{minted}[fontsize=\tiny, gobble=4, frame=single, breaklines=true, breaksymbol={}]{markdown}
    You are a fact-checking assistant operating over structured data. You will be given a natural-language claim and optional external information. You will have access to a SQLite database and may execute arbitrary SQL queries over it using specialized tools.
    
    Your task is to determine whether the claim is "ENTAILED", "CONTRADICTED", or "NOT ENOUGH INFO" based on evidence you obtain from the database. The labels are defined as follows:
    
    - ENTAILED: The claim is supported by the database.
    - CONTRADICTED: The claim is refuted by the database.
    - NOT ENOUGH INFO: The database does not provide sufficient evidence to decide.
    
    Use the available tools to query the database and gather evidence before making a decision. Do not ask the user for clarification or additional information.
    
    You should always start by querying the database for the schema (tables and columns).
    
    Your answer should be in JSON format, adhering to the following schema:
    {
      "properties": {
        "verdict": {
          "description": "Whether the claim is supported, contradicted, or undecidable from the database.",
          "enum": [ "ENTAILED", "CONTRADICTED", "NOT ENOUGH INFO"],
          "title": "Verdict",
          "type": "string"
        },
        "justification": {
          "description": "Brief justification (1-2 sentences) of the verdict.",
          "title": "Justification",
          "type": "string"
        }
      },
      "required": ["verdict", "justification"],
      "title": "ClaimVerdict",
      "type": "object"
    }
    
    
    Output Example 1:
    {
        "verdict": "ENTAILED",
        "justification": "The database shows that the population of France is 67 million, which supports the claim."
    }
    
    Output Example 2:
    {
        "verdict": "CONTRADICTED",
        "justification": "The database indicates that the capital of Germany is Berlin, contradicting the claim."
    }
    
    Output Example 3:
    {
        "verdict": "NOT ENOUGH INFO",
        "justification": "The database does not contain any information about the population of Sacramento."
    }
\end{minted}
\caption{Prompt for all verifier models in the experiments (Section~\ref{sec:setup}). It has been carefully constructed with many iterations and refinements across the different model families. We have tried our best to make the models perform as good as possible. For example, after noticing some models deciding \textit{not} to use tools and hallucinate an answer, we including instructions like \textit{``Use the available tools to query the database and gather evidence before making a decision''} and \textit{``You should always start by querying the database for the schema''}. Furthermore, for a discussion on output format see Appendix~\ref{appendix:structured}. Notably, for the observant readers, there is no explicit explanation of the \textit{tool} the agent has in its disposal because these descriptions are provided by the frameworks internally (most are built this way) by injecting the docstrings and input-output schemas of the tools in the prompt using templates.}
\label{fig:verifier_prompt}
\end{figure}

\FloatBarrier

\begin{table*}[t]
\centering
\tiny
\begin{tabular}{lll}
\toprule
\textbf{Domain} & \textbf{Subdomain} & \textbf{Databases} \\
\midrule
Entertainment & Movies & \texttt{movie}, \texttt{movie\_3}, \texttt{movie\_platform}, \texttt{movies}, \texttt{movies\_4}, \texttt{movielens} \\
 & Music & \texttt{music}, \texttt{music\_tracker}, \texttt{music\_platform\_2} \\
 & TV Shows & \texttt{law\_episode} \\
 & Games & \texttt{video\_games}, \texttt{card\_games} \\
 & Cartoons & \texttt{simpson\_episodes}, \texttt{superhero}, \texttt{disney} \\
\cmidrule{2-3}
Technology & Software & \texttt{talkingdata}, \texttt{codebase\_community}, \texttt{codebase\_comments}, \texttt{social\_media}, \texttt{software\_company} \\
 & IT & \texttt{public\_review\_platform}, \texttt{app\_store} \\
 & Blockchain & \texttt{coinmarketcap} \\
 & Vision & \texttt{image\_and\_language} \\
\cmidrule{2-3}
Education & University & \texttt{student\_club}, \texttt{university}, \texttt{cs\_semester}, \texttt{college\_completion}, \texttt{computer\_student} \\
 & Academia & \texttt{authors}, \texttt{citeseer}, \texttt{book\_publishing\_company} \\
 & Schools & \texttt{california\_schools} \\
 & Language & \texttt{language\_corpus} \\
 & Books & \texttt{books}, \texttt{shakespeare} \\
\cmidrule{2-3}
Health & Healthcare & \texttt{synthea}, \texttt{donor}, \texttt{mental\_health\_survey} \\
 & Medical & \texttt{thrombosis\_prediction} \\
 & Biology & \texttt{genes} \\
 & Chemistry & \texttt{toxicology} \\
\cmidrule{2-3}
Economy & Finance & \texttt{student\_loan}, \texttt{debit\_card\_specializing} \\
 & World Economies & \texttt{world\_development\_indicators} \\
 & Retail & \texttt{retail\_complains}, \texttt{sales}, \texttt{superstore}, \texttt{car\_retails}, \texttt{regional\_sales}, \texttt{retails}, \texttt{retail\_world}, \texttt{works\_cycles} \\
 & Banking & \texttt{financial} \\
\cmidrule{2-3}
Transportation & Transit Systems & \texttt{bike\_share\_1}, \texttt{shipping}, \texttt{trains}, \texttt{cars} \\
 & Airport & \texttt{airline} \\
\cmidrule{2-3}
Gastronomy & Food & \texttt{food\_inspection\_2}, \texttt{beer\_factory}, \texttt{food\_inspection}, \texttt{cookbook}, \texttt{craftbeer} \\
 & Restaurant & \texttt{restaurant}, \texttt{menu} \\
\cmidrule{2-3}
Governance & Crime & \texttt{chicago\_crime}, \texttt{shooting} \\
 & Law & \texttt{legislator} \\
\cmidrule{2-3}
Environment & Weather & \texttt{sales\_in\_weather} \\
 & Geography & \texttt{mondial\_geo}, \texttt{address}, \texttt{world} \\
\cmidrule{2-3}
Labor & Human Resources & \texttt{human\_resources} \\
\cmidrule{2-3}
Sports & Basketball & \texttt{professional\_basketball} \\
 & Olympics & \texttt{olympics} \\
 & Hockey & \texttt{ice\_hockey\_draft}, \texttt{hockey} \\
 & F1 & \texttt{formula\_1} \\
 & Soccer & \texttt{european\_football\_1}, \texttt{european\_football\_2}, \texttt{soccer\_2016} \\
\cmidrule{2-3}
\bottomrule
\end{tabular}
\caption{Domain taxonomy with representative databases (80 total across 11 domains and 36 subdomains).}
\label{tab:full_domain_taxonomy}
\end{table*}

\end{document}